\documentclass[12pt,twocolumn,notitlepage]{article}

\usepackage{graphicx}
\DeclareGraphicsExtensions{.eps}

\usepackage[utf8]{inputenc}

\usepackage{subcaption}

\usepackage{multirow}
\usepackage[table, dvipsnames]{xcolor}

\usepackage{array}
\newcolumntype{L}[1]{>{\raggedright\let\newline\\\arraybackslash\hspace{0pt}}m{#1}}
\newcolumntype{C}[1]{>{\centering\let\newline\\\arraybackslash\hspace{0pt}}m{#1}}
\newcolumntype{R}[1]{>{\raggedleft\let\newline\\\arraybackslash\hspace{0pt}}m{#1}}

\usepackage{needspace}
\usepackage{makeidx}  
\usepackage{changes}
\usepackage{todonotes} 

\makeatletter
\setremarkmarkup{\todo[color=Changes@Color#1!10,size=\scriptsize]{#1: #2}}
\setremarkmarkup{\todo[backgroundcolor=Changes@Color#1!20!white,linecolor=Changes@Color#1!30!white,size=\scriptsize\sffamily]{\@nameuse{Changes@AuthorName#1}\\#2}}
\makeatother

\newcommand*{\helvetica}{\fontfamily{phv}\selectfont\scriptsize}

\definecolor{blueNote}{HTML}{2B619D}
\definechangesauthor[color=blueNote,name=\textbf{Alejandro}]{AC}
\definechangesauthor[color=red,name=\textbf{Mariella}]{MD}

\begin{document}

\title{Batch-based Activity Recognition from Egocentric Photo-Streams Revisited}

\author{Alejandro Cartas\and Juan Mar\'in\and Petia Radeva \and Mariella Dimiccoli}

\maketitle

\begin{abstract}
Wearable cameras can gather large a\-mounts of image data that provide rich visual information about the daily activities of the wearer. Motivated by the large number of health applications that could be enabled by the automatic recognition of daily activities, such as lifestyle characterization for habit improvement, context-aware personal assistance and tele-rehabilitation services, we propose a system to classify 21 daily activities from photo-streams acquired by a wearable photo-camera. Our approach combines the advantages of a Late Fusion Ensemble strategy relying on convolutional neural networks at image level with the ability of recurrent neural networks to account for the temporal evolution of high level features in photo-streams without relying on event boundaries. The proposed batch-based approach achieved an overall accuracy of 89.85\%, outperforming state of the art end-to-end methodologies. These results were achieved on a dataset consists of 44,902 egocentric pictures from three persons captured during 26 days in average.
\end{abstract}

\section{Introduction}

During the last decade there has been a growing interest in analyzing human activities from sensory data \cite{mukhopadhyay2015wearable}. More recently, the introduction of wearable cameras opened the unique opportunity to capture richer contextual information than using only traditional sensors \cite{Nguyen2016}. Lifelogging cameras, as a particular case of wearable cameras, are the only tools that allow to capture image data during several days thanks to their very low frame rate (2fpm) without recharging its battery. Consequently, they are best suited to monitor the Activities of Daily Living (ADLs) of the wearer that include, but are not limited to, the activities that an independent person performs on daily basis for living at home or in a community \cite{martin2015design}. The monitoring of these activities has several applications in health-related research including active aging monitoring, tele-rehabilitation, frailty prevention and stroke-survivor monitoring among others \cite{martin2015design,schussler2016potentially}.

\begin{figure*}[!t]
\begin{center}
\newcommand\imgscale{0.055}
\begin{minipage}{\textwidth}
\centering
\rotatebox{90}{{\hspace{0.5cm}\helvetica Biking}}%
\includegraphics[scale=\imgscale]{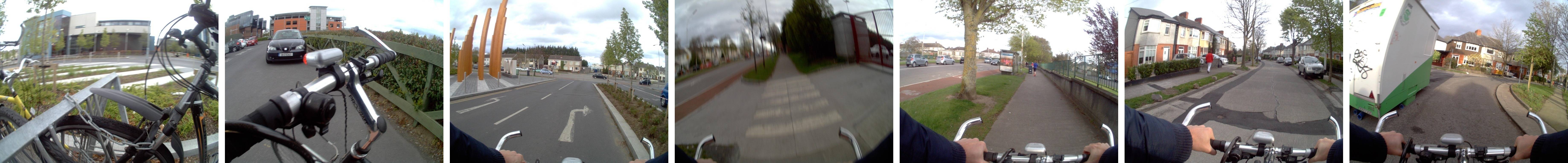}%
\end{minipage}

\begin{minipage}{\textwidth}
\centering
\rotatebox{90}{{\hspace{-0.1cm}\helvetica Walking outdoor}}%
\includegraphics[scale=\imgscale]{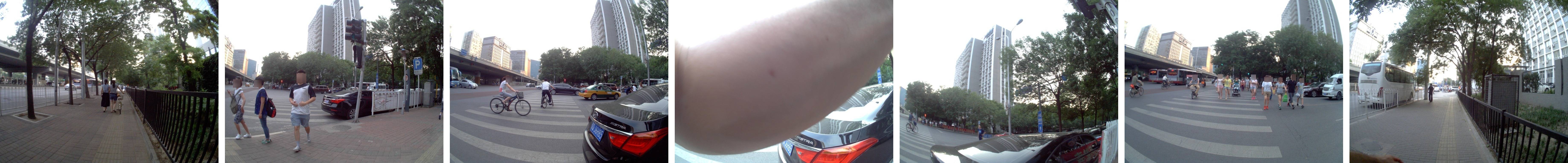}%
\end{minipage}

\begin{minipage}{\textwidth}
\centering
\rotatebox{90}{{\hspace{0.5cm}\helvetica TV}}%
\includegraphics[scale=\imgscale]{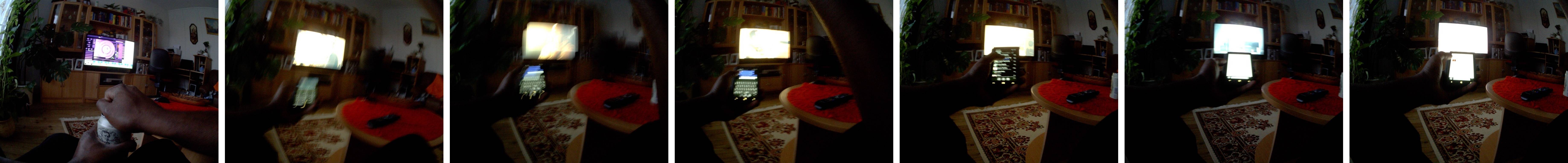}%
\end{minipage}

\begin{minipage}{\textwidth}
\centering
\rotatebox{90}{{\hspace{0.25cm}\helvetica Reading}}%
\includegraphics[scale=\imgscale]{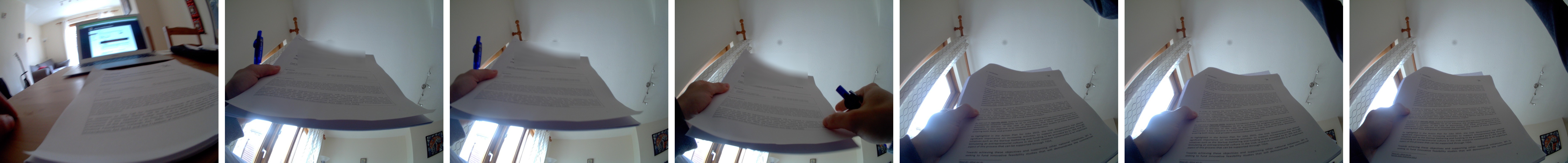}%
\end{minipage}

\caption[]{Examples of consecutive frames captured while the user was performing four activities: Biking, Walking outdoor, TV, and Reading.}
\label{fig:examplesFrameSequences}
\end{center}
\end{figure*}

However, activity recognition from first-person (egocentric) photo-streams has received relatively little attention in the literature \cite{castro2015predicting,cartas2017recognizing,oliveira2017leveraging,cartas2017batch}. One of its major challenges is that photo-streams are characterized by a very low frame-rate, and consequently useful important features such as optical flow cannot be reliably estimated. With the only exception of \cite{cartas2017batch}, all existing methods treat photo-streams as a set of unrelated images, neglecting the fact that while the user is performing a given activity such as \textit{cooking}, the temporal coherence of environment concepts such as \textit{food}, \textit{kitchen}, \textit{pan} is preserved despite of suddenly image appearance changes. This can be easily appreciated in Fig. \ref{fig:examplesFrameSequences}, where examples of consecutive frames are shown while the user is performing different activities, including very dynamic activities such as \textit{biking} and \textit{walking outdoor}. Additionally, some images can contain poor information that is difficult to interpret without considering it within its temporal image context. For instance in the 4th image of the activity \textit{walking outdoor} in Fig. \ref{fig:examplesFrameSequences}, it would be impossible to assign the correct label without looking at neighboring images.

In this paper, we propose a new method for activity recognition from photo-streams that is based on our previous work \cite{cartas2017recognizing}, but extends it by taking into account the temporal coherence of high level features. More specifically, in \cite{cartas2017recognizing} we applied a Late Fusion Ensemble (LFE) method that merges through a random decision forest the activity probabilities extracted by a Convolutional Neural Network (CNN) with contextual information embedded in a fully connected layer of the CNN. In this work, we integrate a random decision forest within a temporal framework implemented in terms of a Long Short Term Memory (LSTM) recurrent neural network that processes overlapping batches of fixed size and learns the temporal evolution of high level features.

With respect to \cite{cartas2017recognizing}, our contributions in this paper are as follows:
\begin{itemize}
\item we straighten our originally proposed pipeline by integrating for the first time LFE in a framework based on the temporal coherence of high-level features in low temporal resolution photo-streams instead of treating images as non-temporally related,
\item we extend the annotated dataset from $\approx$18K images to about $\approx$45K and made the annotations publicly available \footnote{The annotations are publicly available at \textit{https://www.github.com/gorayni/egocentric\_photostreams}.}, we call it the UB Extended Annotations (UBEA) dataset,
\item we extend significantly the comparisons with state of the art techniques.
\item we provide a more extensive validation of the method previously proposed by us in \cite{cartas2017recognizing}

\end{itemize}

The reminder of the paper is organized as follows: in section \ref{sec:SoA}, we discuss related work; in section \ref{sec:activityClassification}, we detail the proposed approach. In section \ref{sec:dataset}, we introduce the dataset used in the experiments. We detail the methodology we followed for conducting our experiments and the different combinations of networks and layers we used in section \ref{sec:training}, which details the training settings. The results we obtained are discussed in section \ref{sec:results}. Finally, we present our conclusions and final remarks in section \ref{sec:conclusion}.

\section{Related work}
\label{sec:SoA}

\textbf{Activity recognition from egocentric video}. Recently, activity recognition from egocentric videos has become an active area of research, specially using high-temporal frame rate videos. Fathi et al. \cite{fathi2011understanding} proposed a probabilistic model that maps activities into a set of multiple actions, and each action is modeled per frame as a spatio-temporal relationship between the hands and the objects involved on it. They further extend their work \cite{fathi2012learning} by proposing another probabilistic generative model that incorporates the gaze features and that models an action as a sequence of frames. Pirsiavash and Ramanan \cite{pirsiavash2012detecting} introduced a dataset of 18 egocentric actions of daily activities performed by 20 persons in unscripted videos. On this dataset, they presented a temporal pyramid to encode spatio-temporal features along with detected active objects knowledge. These temporal pyramids are the input of support vector machines trained for action recognition. More recently, Ma et al. \cite{Ma_2016_CVPR} proposed a twin stream CNN architecture for activity recognition from videos. One of the streams is used for recognizing the appearance of an object based on a hand segmentation and a region of interest. The other stream recognizes the action using an optical flow sequence. In order to recognize activities, both streams are joined and the last layers are fine-tuned.

\begin{figure*}[!h]
\begin{center}
\includegraphics[scale=.32]{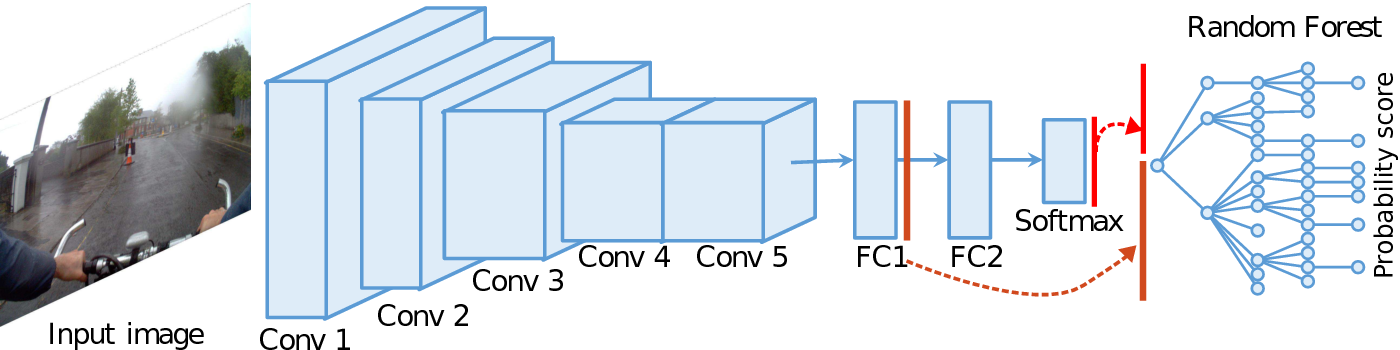}
\caption[]{Activity recognition at image level. After fine-tuning a CNN, we combine the softmax probabilities and a fully connected layer into a single vector. Later on, we train a random forest using this vector as its input.}
\label{fig:overviewFirstMethod}
\end{center}
\end{figure*}

The method proposed by Singh et al. \cite{Singh_2016_CVPR} first extracts hand masks, head motion, and a saliency map from a sequence of input images. Then, it uses these features as the input of a neural network architecture that combines four streams. The first two streams capture temporal egocentric features by processing a video segment containing binary hand-arm masks. The third stream processes spatial information by using as input the full RGB frame. The last stream handles motion taken from stacked optical flow encoded in a saliency map.

\textbf{Activity recognition from egocentric photo-streams}.%
In contrast to all above mentioned methods that use egocentric videos, activity classification from low-temporal frame rate egocentric photo-streams captured by lifelogging devices has received comparatively little attention in the literature \cite{bolanos2017toward}. This may be due partially to the lack of available benchmarks, and partially to the fact that photo-streams provide less  contextual action information with respect to video data since images are taken at periodic intervals of 20 or 30 seconds. Due to the lack of temporal coherence, motion features that are the most largely exploited in egocentric videos, cannot be reliably estimated. The pioneering work of Castro et. al. \cite{castro2015predicting} proposed a LFE method that combines, through a random decision forest, the classification probabilities of a CNN with time and global features, namely color histogram, from the input image, to classify images into 19 different activity categories. Their method has been tested on a nonpublic lifelogging dataset made of 40,103 egocentric images captured by a single person during a 6 months period. Since the user activities are performed almost daily at the same time and in the same environment, time and global image features such as color convey useful information for describing the activities in the context of the same wearer. However, the method can hardly generalize to multiple users, since the network needs to adapt or be trained again with respect to the contextual information for each new user. For instance, two distinct persons might have different daily routines, depending on their job, hobbies, age, etc. and, consequently, the system needs to deal with an increased intra-class variability. For instance the ``work'' activity for different persons could have a very difference appearance.

To address the problem of generalizing the LFE method to multiple users, Cartas et al. \cite{cartas2017recognizing} used the outputs of different layers of a CNN as contextual information instead of using color and time information that are too much tied to a single user context. The authors tested their approach on a dataset acquired by three different users having different lifestyles. Instead of focusing on contextual information, Oliveira et al. \cite{oliveira2017leveraging} used a gradient boosting machine approach to retrieve activities based on their estimated relations with objects in the scene. Their method was tested on an extended version of the dataset used by Cartas el al. \cite{cartas2017recognizing}, showing promising results.

While all the approaches described so far treated photo-streams as a set of temporally unrelated images, more recently Cartas et al. \cite{cartas2017batch} proposed an end-to-end approach that takes into account the temporal coherence of photo-streams. Their proposed architecture consists of Long Short Term Memory units on the top of a CNN for each frame that is trained by processing the photo-streams using batches of fixed size. Experimental results over the same dataset used in \cite{oliveira2017leveraging}, have shown a $6\%$ improvement in accuracy with respect to the VGG-16 baseline.

Inspired by these results, we propose here a new approach that exploits the temporal evolution of high-order features to improve the results of the LTE method proposed in \cite{cartas2017recognizing}.

\section{Egocentric activity classification}
\label{sec:activityClassification}

Our base activity classification method is an ensemble classifier composed of a CNN and a random forest, as illustrated on Fig. \ref{fig:overviewFirstMethod}, that acts at image level. Specifically, the random forest takes as input one or more concatenated output vectors from the final layers of a CNN. Depending on the CNN architecture, the input for the random forest can be extracted from the last convolutional layer, a fully-Connected (FC) layer, or the softmax layer. For instance, a random forest that takes as input the output of the softmax and a FC layer is shown on Fig. \ref{fig:overviewFirstMethod}. The training of the ensemble is a two-step process. In the first step, the CNN is fine-tuned. In the second step, a random forest is trained over the output vectors of the CNN. The insight underlying this method is that a fully connected layer provides global contextual information which helps to generalize the features characterizing a given activity among different users.

\begin{figure*}[t]
  \centering
  \begin{subfigure}[b]{0.4\textwidth}
      \includegraphics[scale=0.52]{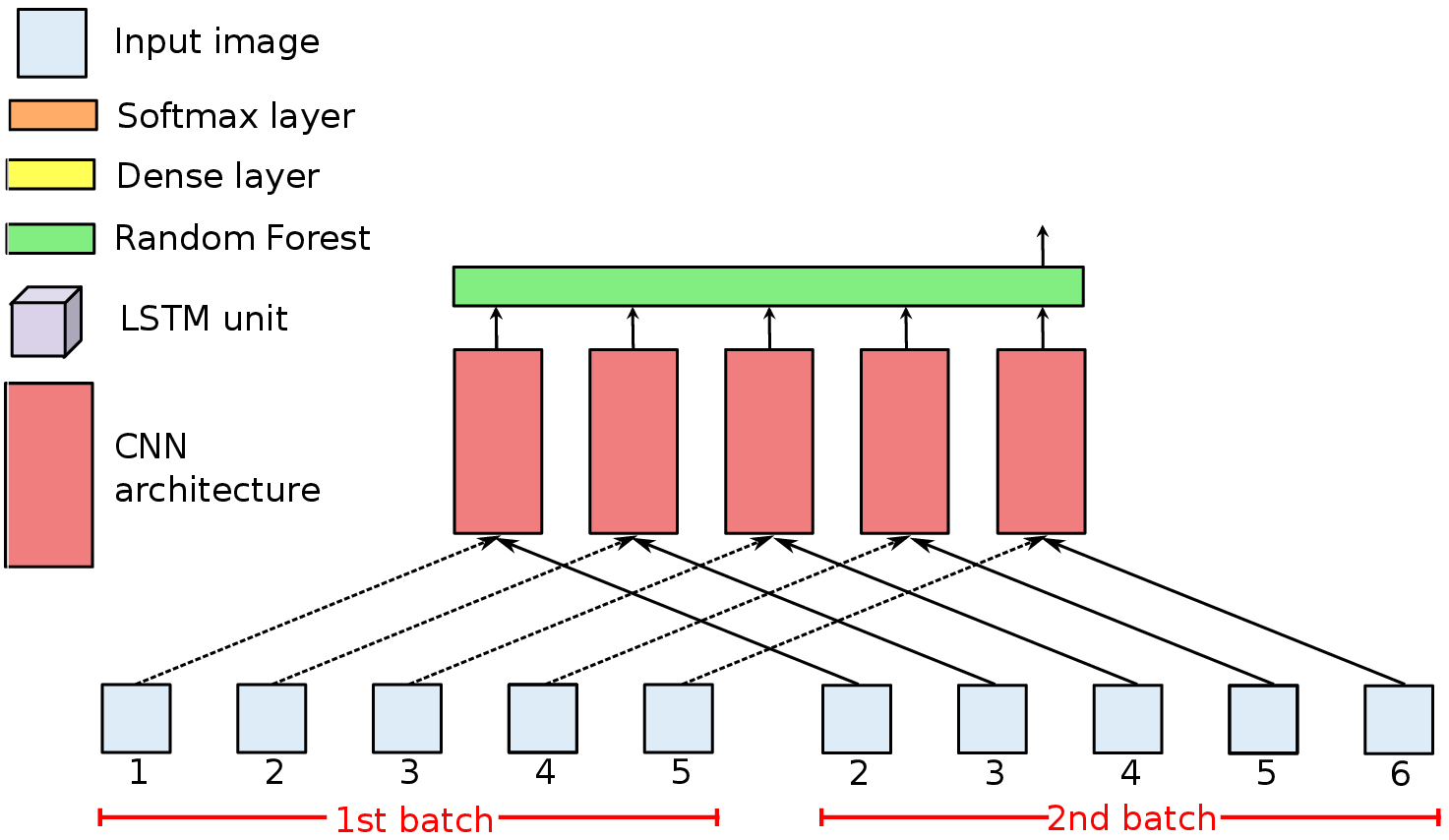}
      \caption[]{}
      \label{fig:temporal_many2one_RF}
  \end{subfigure}\hspace{1.5cm}
  \begin{subfigure}[b]{0.4\textwidth}
      \includegraphics[scale=0.52]{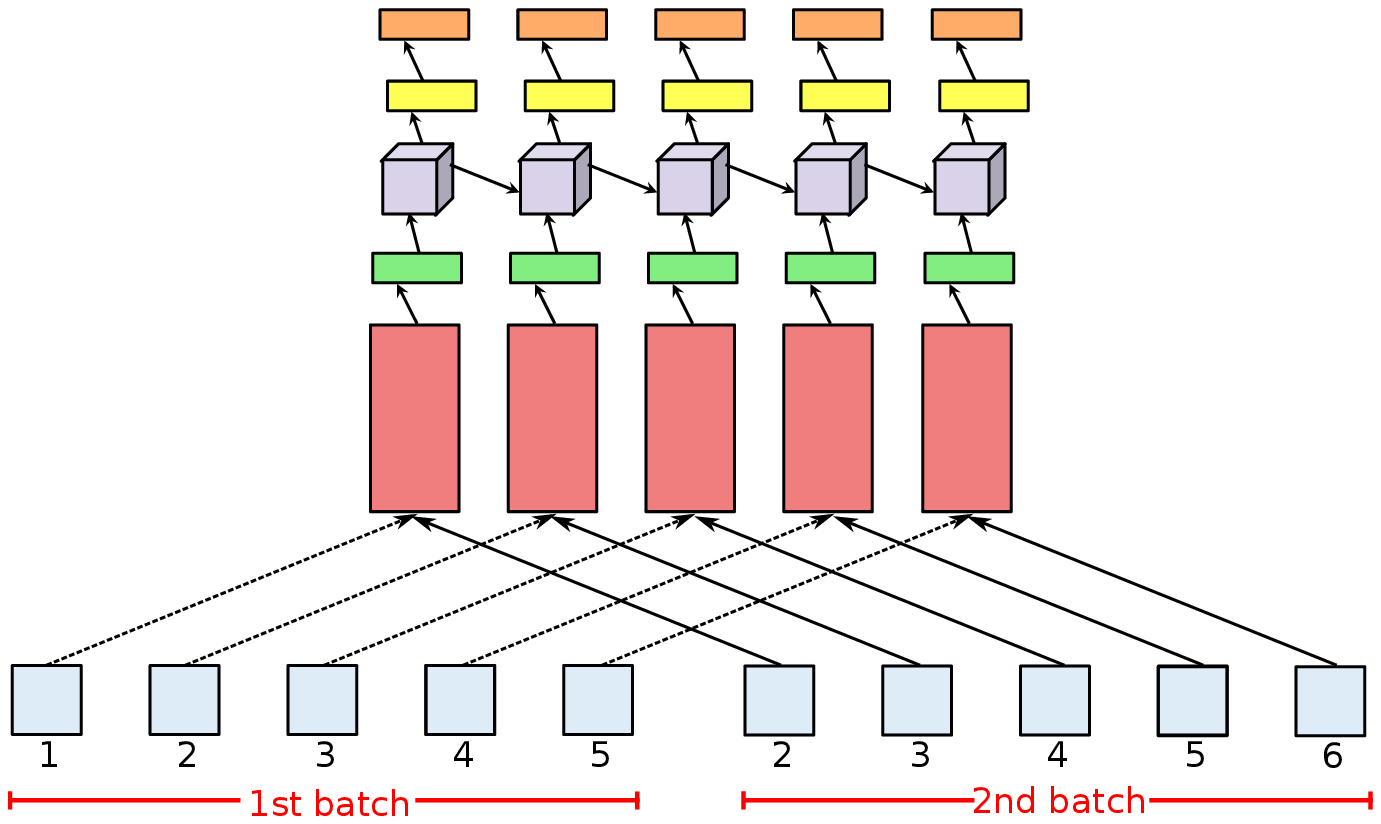}
      \caption[]{}      
      \label{fig:temporal_RF_LSTM}
  \end{subfigure}
\caption[]{Activity recognition using temporal contextual information. With the activity recognition scores obtained as shown in Fig. \ref{fig:overviewFirstMethod}, in a) we train a random forest by feeding it with overlapping batches of fixed size and a single label is assigned to each image of the batch; b) we train a many-to-many LSTM, feeding it in the same way than in (a).}
\label{fig:temporalMethods}
\end{figure*}

With the goal of improving classification performance while keeping the advantage of the ensemble classifier, we extended the ensemble architecture to take into account the temporal information from neighbor frames. Firstly, we considered overlapping batches of fixed size and we used a classifier to make a single prediction for each batch. From an implementation point of view, this is equivalent to have a many-to-one predictor since to all frames of the batch is assigned a single label. More specifically, we employed a random forest as classifier that takes as input the concatenation of the outputs of the first fully-connected layer of the CNN of $n$ consecutive frames as shown in Fig. \ref{fig:temporal_RF_LSTM}. This choice was motivated by the fact that it is less prone to overfitting, does not expect linear features and has a small number of hyperparameters. However, since the Random Forest does not have an internal memory state,  it does no take into account explicitly the temporal evolution of high level features, but the set of temporal predictions as a whole.

Secondly,  encouraged by the promising results achieved in \cite{cartas2017batch}, showing that it is possible to capture the temporal evolution of high-level features extracted with a CNN without knowing event boundaries, and by the success of LSTM in video classification tasks \cite{yue2015beyond,donahue2015}, we propose the architecture showed in Fig. \ref{fig:temporal_RF_LSTM}. More specifically, we introduce a many-to-many LSTM on the top of the ensemble previously described. Previous attempts to employ LSTM for modeling the temporal evolution of features over time have led to end-to-end trainable architectures \cite{yue2015beyond,donahue2015,cartas2017batch}. Similarly to the architecture proposed in Fig. \ref{fig:temporal_many2one_RF}, the one proposed in Fig. \ref{fig:temporal_RF_LSTM} is not end-to-end: the input of the LSTM is the probability classification score from the ensemble. The training is performed in two phases. We first train an ensemble as stated in the previous section. In order to reduce the computational cost in the second phase, we store the classification scores of each training image. During the second phase, the LSTM unit is trained using the classification scores of $n$ consecutive frames. As a way of performing data augmentation, the training batches are sampled using a sliding window. For example, Fig. \ref{fig:temporal_RF_LSTM} shows two consecutive batches of 5 frames that are sampled.

\section{Validation}
\label{sec:experiments}

The main objective of the experiments performed in this work is to prove that temporal contextual information combined with a CNN LFE improves the state of the art activity classification accuracy on egocentric photo-streams. To this goal, we performed two classes of experiments. Specifically, the goal of the first class of experiments was to determine the ensemble with the best combination of layers for performance improvement based on the proposed approach, firstly proposed in \cite{cartas2017recognizing}, illustrated in Fig. \ref{fig:overviewFirstMethod}. In these experiments, we used three networks as the base of our ensembles, namely the VGG-16 \cite{Simonyan14c}, InceptionV3 \cite{Szegedy_2016_CVPR}, and ResNet 50 \cite{He_2016_CVPR}. The goal of the second class of experiments was to determine the best approach to exploit temporal contextual information from consecutive frames. We used the same training setting and CNN as in \cite{cartas2017batch} to compare and evaluate the approaches depicted in Fig. \ref{fig:temporalMethods} directly to \cite{cartas2017batch}, that used the same setting, and other state of the art methods.

We describe the dataset in section \ref{sec:dataset} and detail the ensembles training in section \ref{sec:training}. We then present the experimental results on activity recognition in section \ref{sec:results}.

\subsection{Dataset}
\label{sec:dataset}

In our experiments, we used a subset of images from the NTCIR-12 dataset \cite{gurrin2016NTCIR}. This dataset consists of 89,593 egocentric pictures belonging to three persons and acquired with an OMG Autograph camera that captured two pictures per minute. Each user worn this camera in a period about three weeks, totaling 79 days. 

\begin{figure*}[!t]
\begin{center}
\newcommand\rowspace{0.12cm}
\newcommand\imgscale{0.06}
\newcommand\columnProportion{0.2815}
\begin{center}
\begin{minipage}{\columnProportion\columnwidth}
\centering
{\helvetica Attending a Seminar}
\includegraphics[scale=\imgscale]{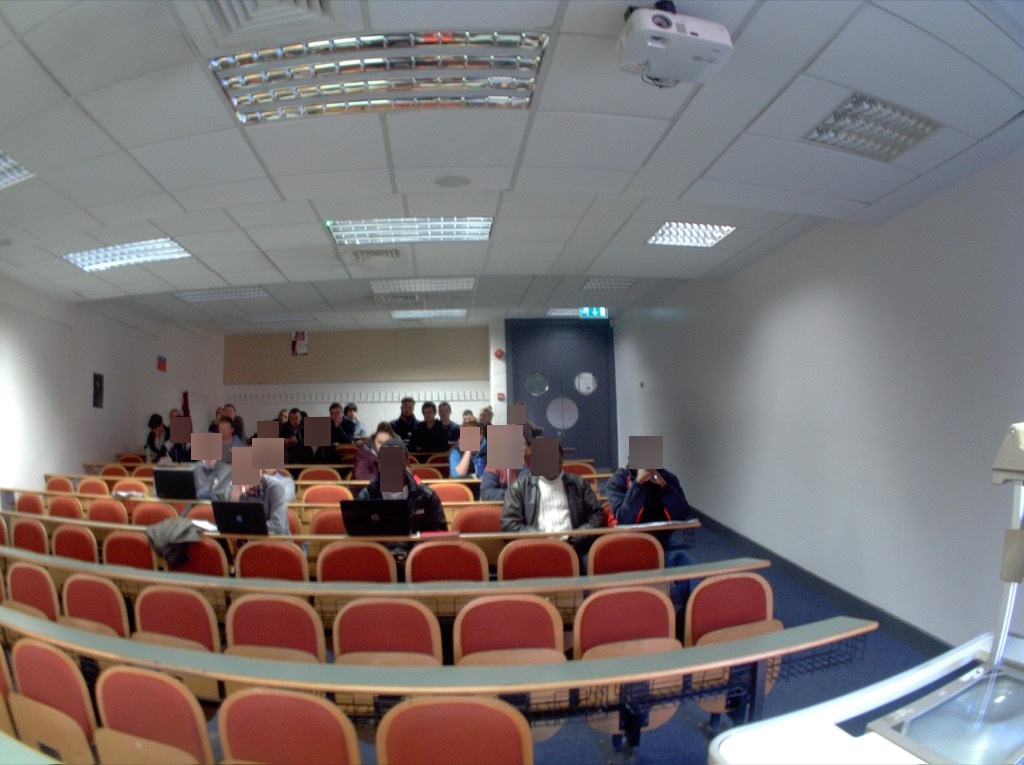}
\end{minipage}%
\begin{minipage}{\columnProportion\columnwidth}
\centering
{\helvetica Biking}
\includegraphics[scale=\imgscale]{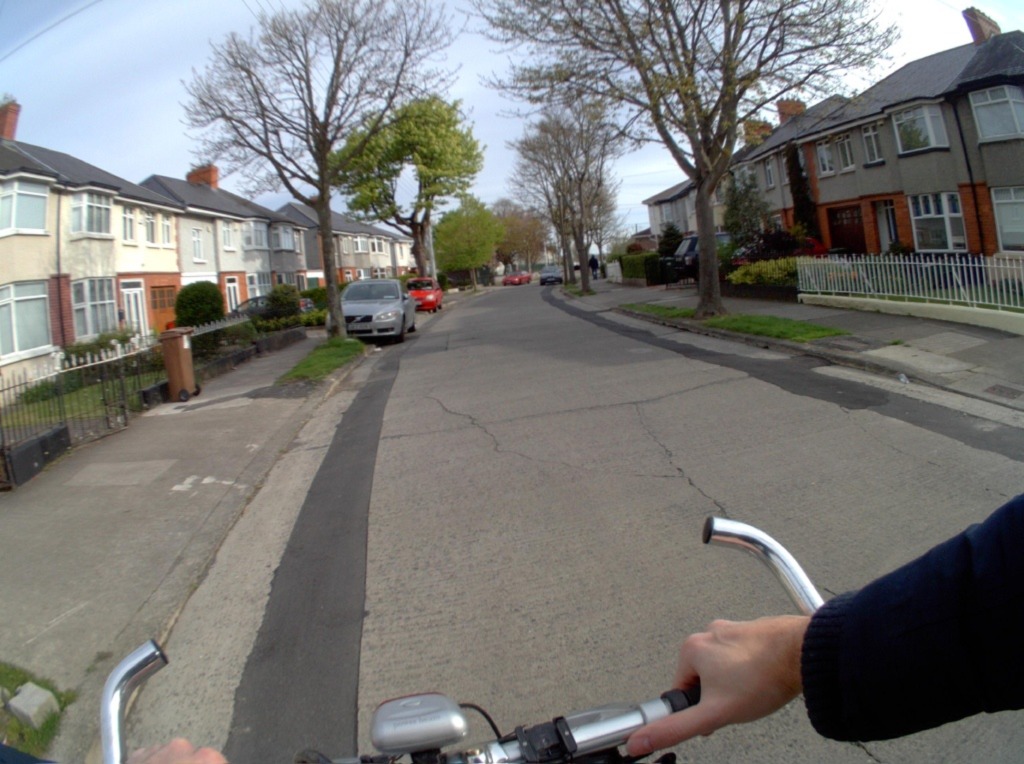}
\end{minipage}%
\begin{minipage}{\columnProportion\columnwidth}
\centering
{\helvetica Driving}
\includegraphics[scale=\imgscale]{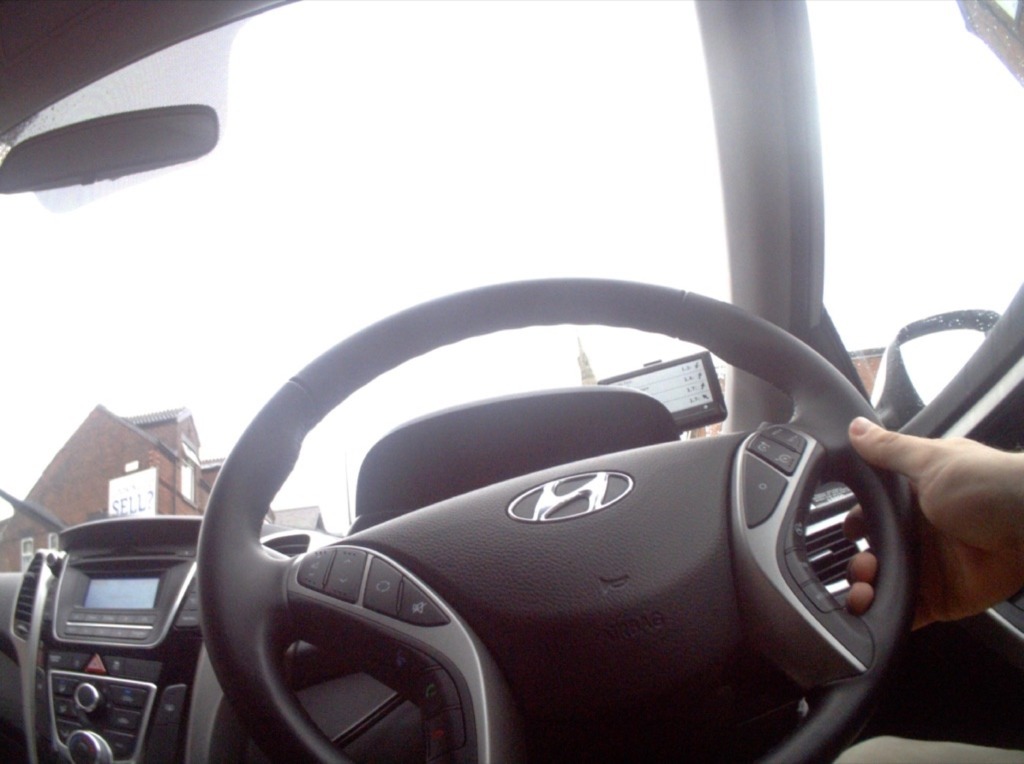}
\end{minipage}%
\begin{minipage}{\columnProportion\columnwidth}
\centering
{\helvetica Talking}
\includegraphics[scale=\imgscale]{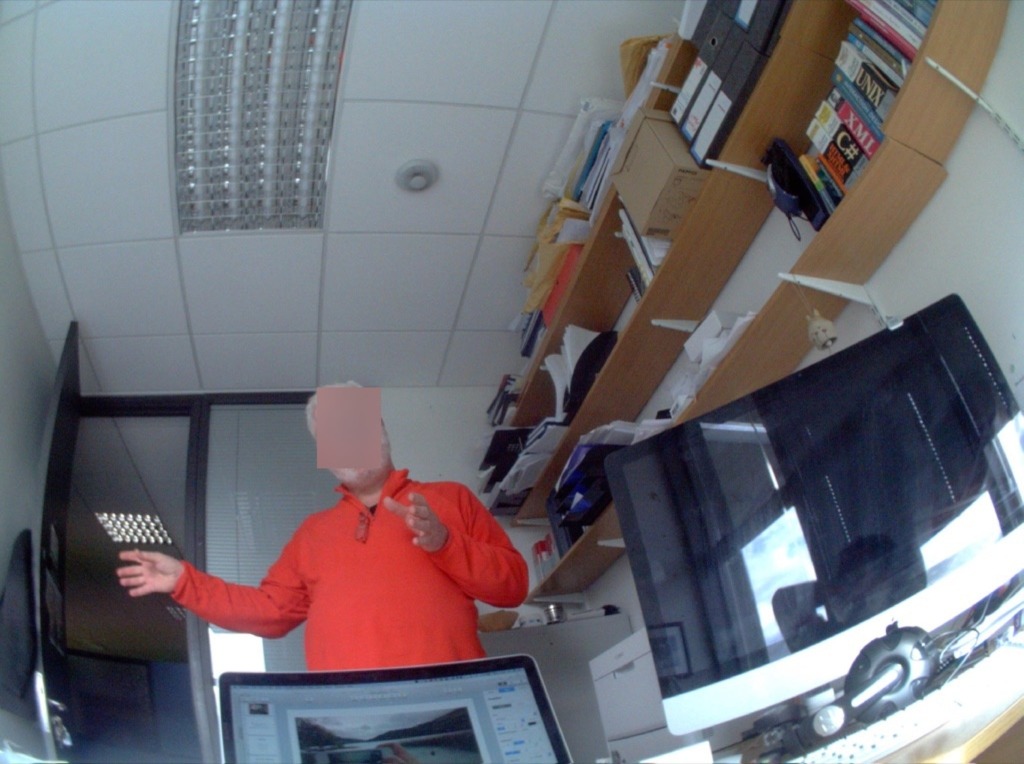}
\end{minipage}%
\begin{minipage}{\columnProportion\columnwidth}
\centering
{\helvetica Plane}
\includegraphics[scale=\imgscale]{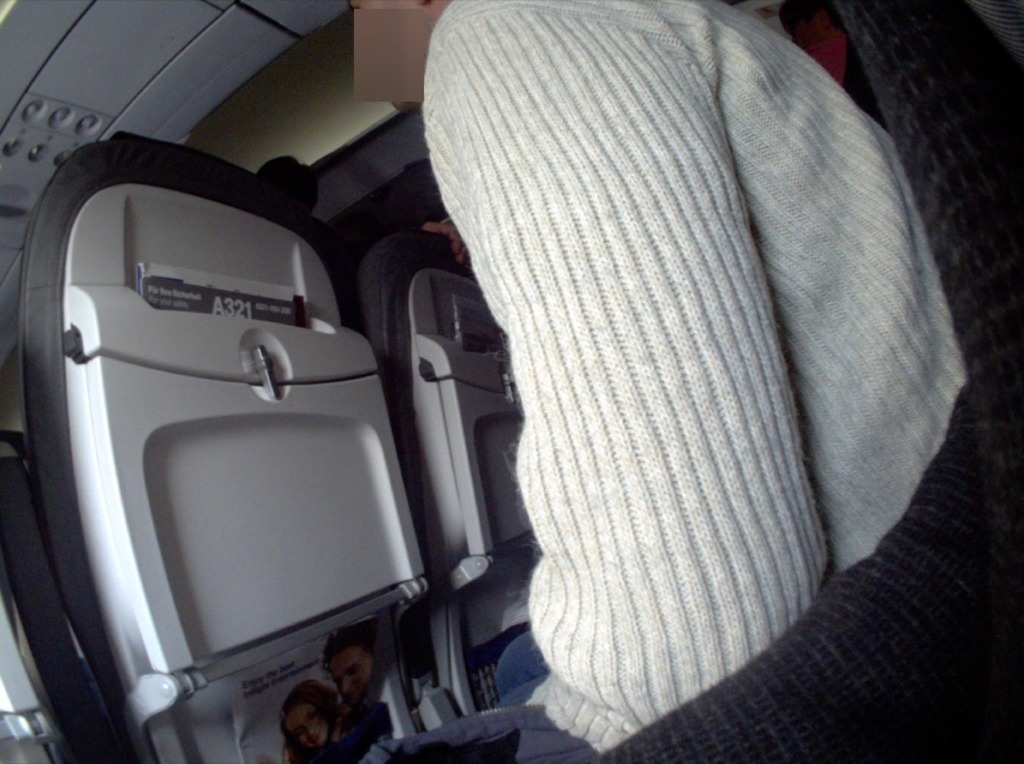}
\end{minipage}%
\begin{minipage}{\columnProportion\columnwidth}
\centering
{\helvetica Shopping}
\includegraphics[scale=\imgscale]{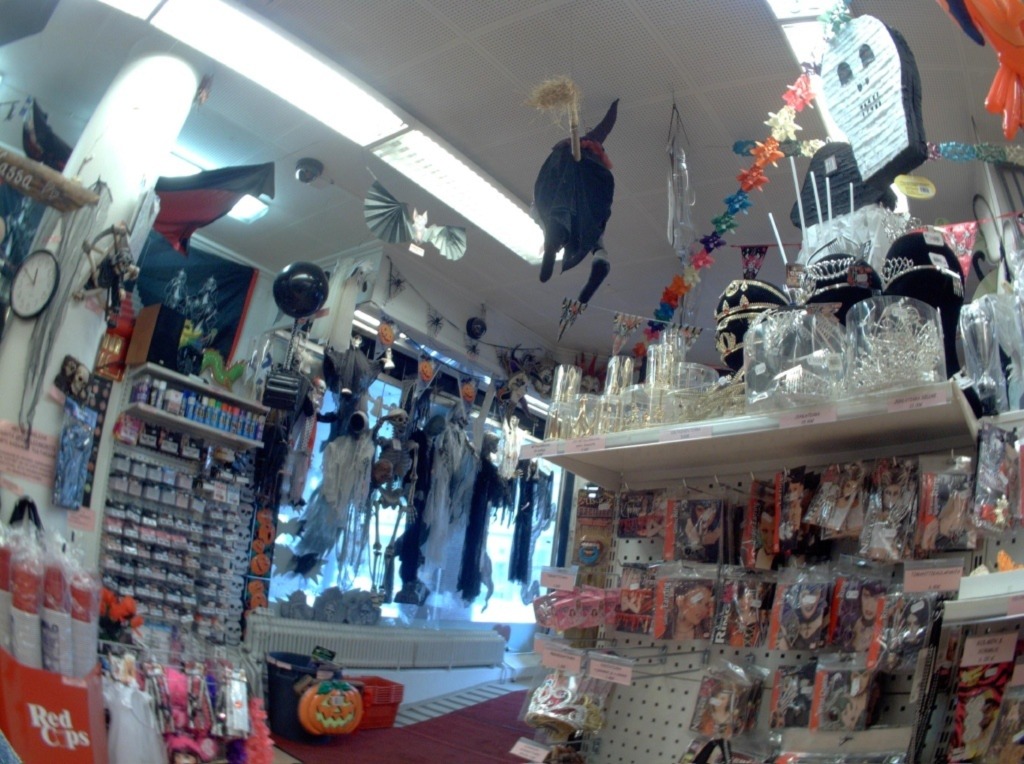}
\end{minipage}
\begin{minipage}{\columnProportion\columnwidth}
\centering
{\helvetica Walking Outdoor}
\includegraphics[scale=\imgscale]{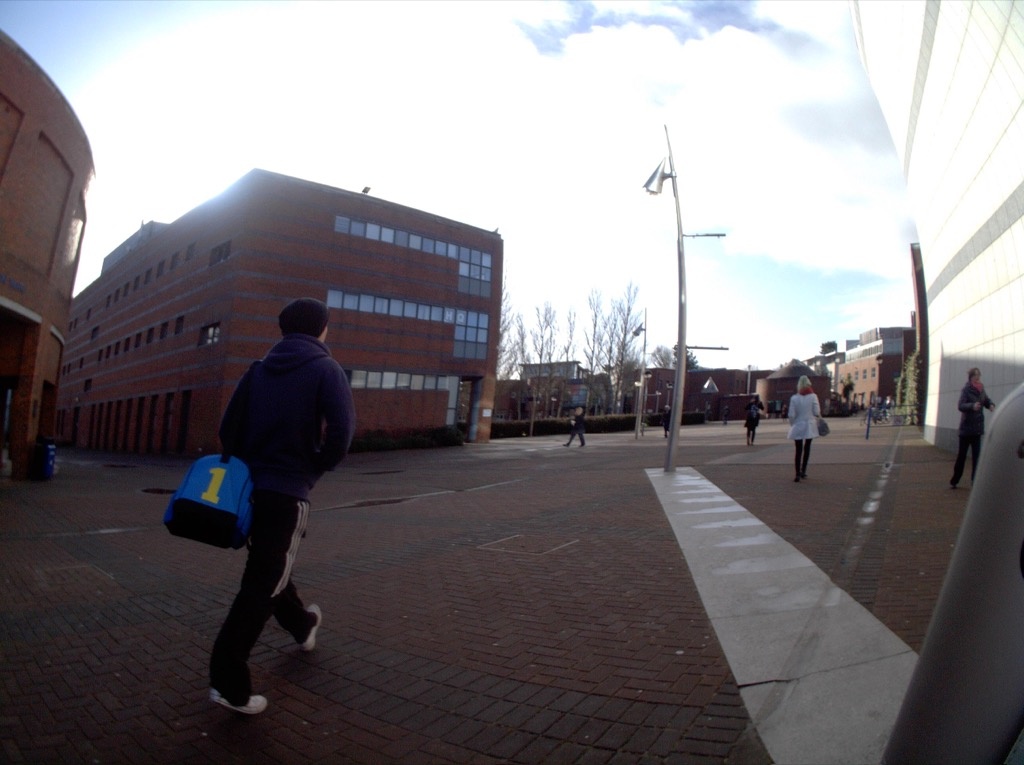}
\end{minipage}\vspace{\rowspace}%

\begin{minipage}{\columnProportion\columnwidth}
\centering
{\helvetica Meeting}
\includegraphics[scale=\imgscale]{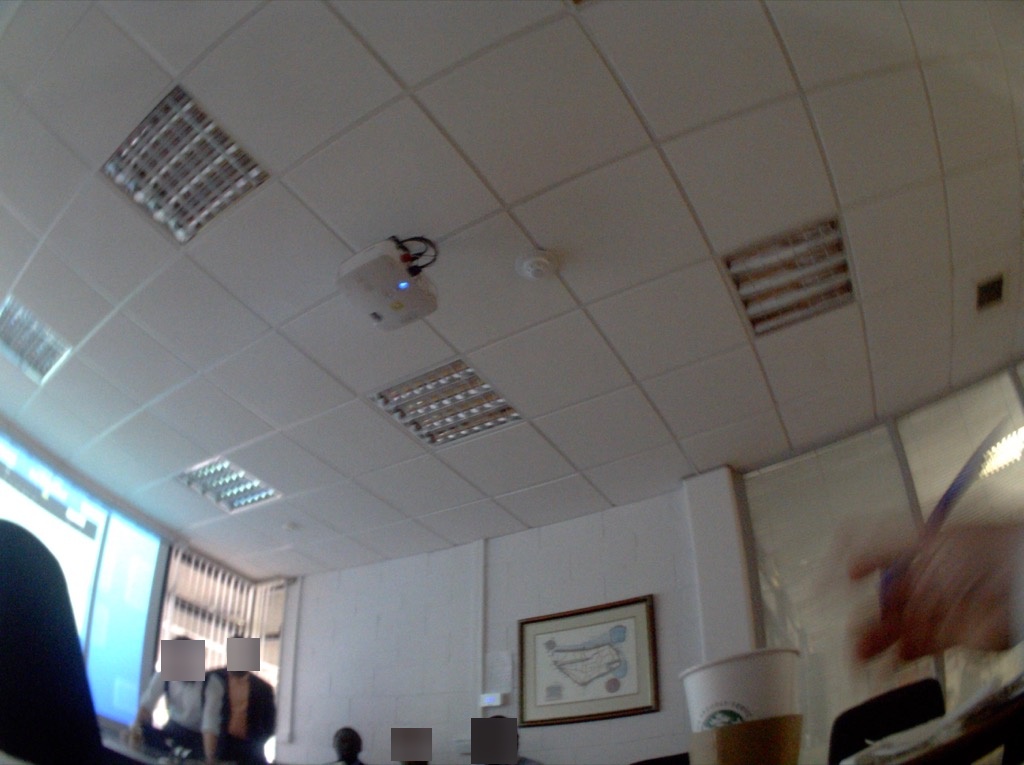}
\end{minipage}%
\begin{minipage}{\columnProportion\columnwidth}
\centering
{\helvetica Cooking}
\includegraphics[scale=\imgscale]{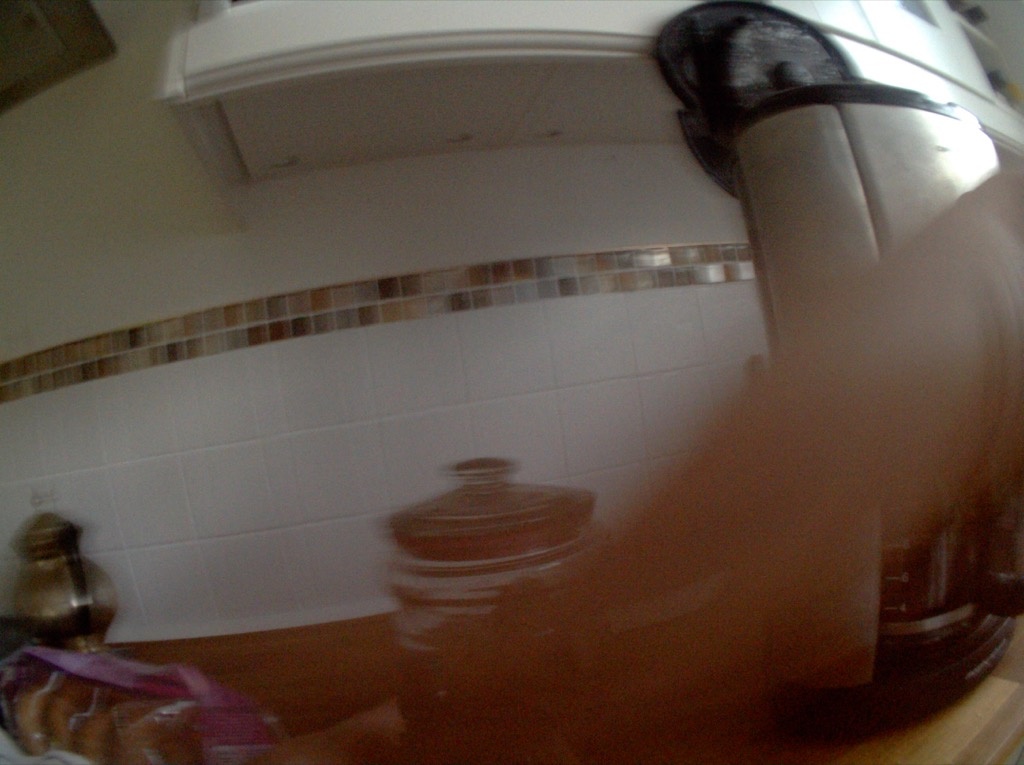}
\end{minipage}%
\begin{minipage}{\columnProportion\columnwidth}
\centering
{\helvetica Cleaning and chores}
\includegraphics[scale=\imgscale]{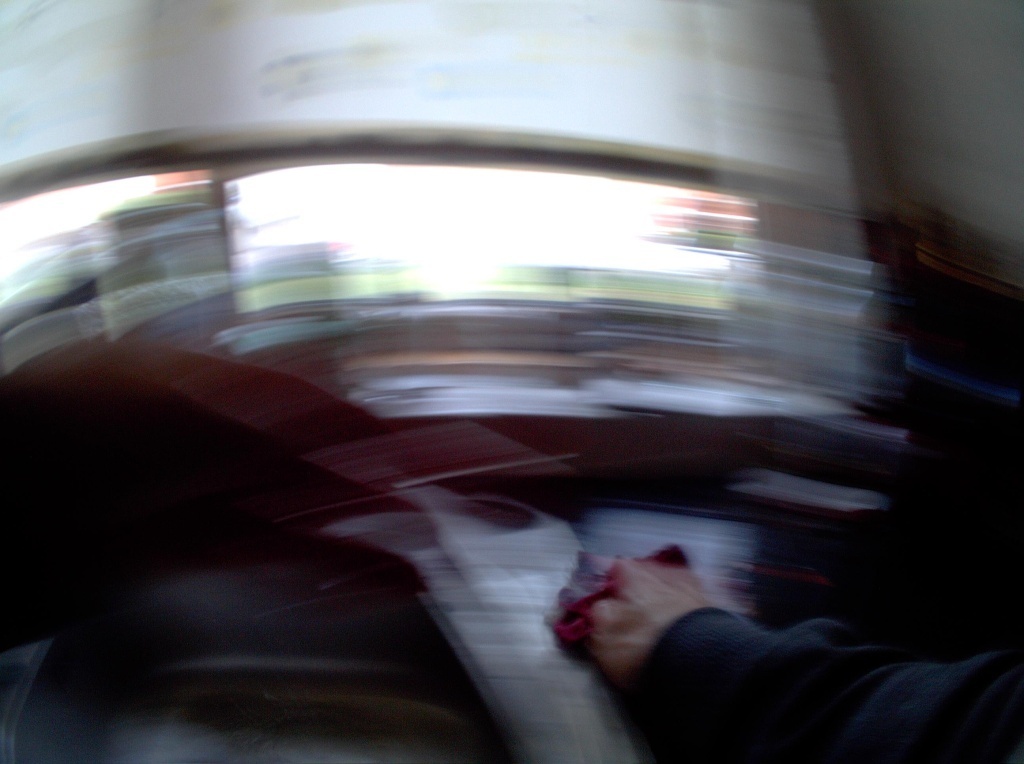}
\end{minipage}%
\begin{minipage}{\columnProportion\columnwidth}
\centering
{\helvetica TV}
\includegraphics[scale=\imgscale]{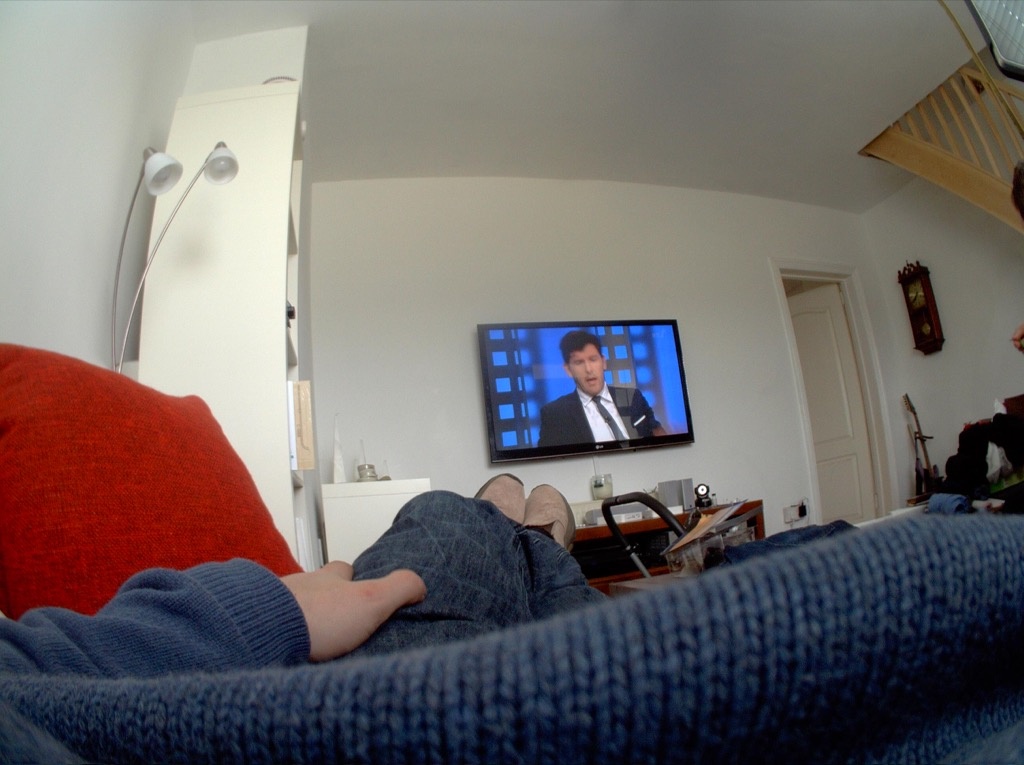}
\end{minipage}
\begin{minipage}{\columnProportion\columnwidth}
\centering
{\helvetica Drinking together}
\includegraphics[scale=\imgscale]{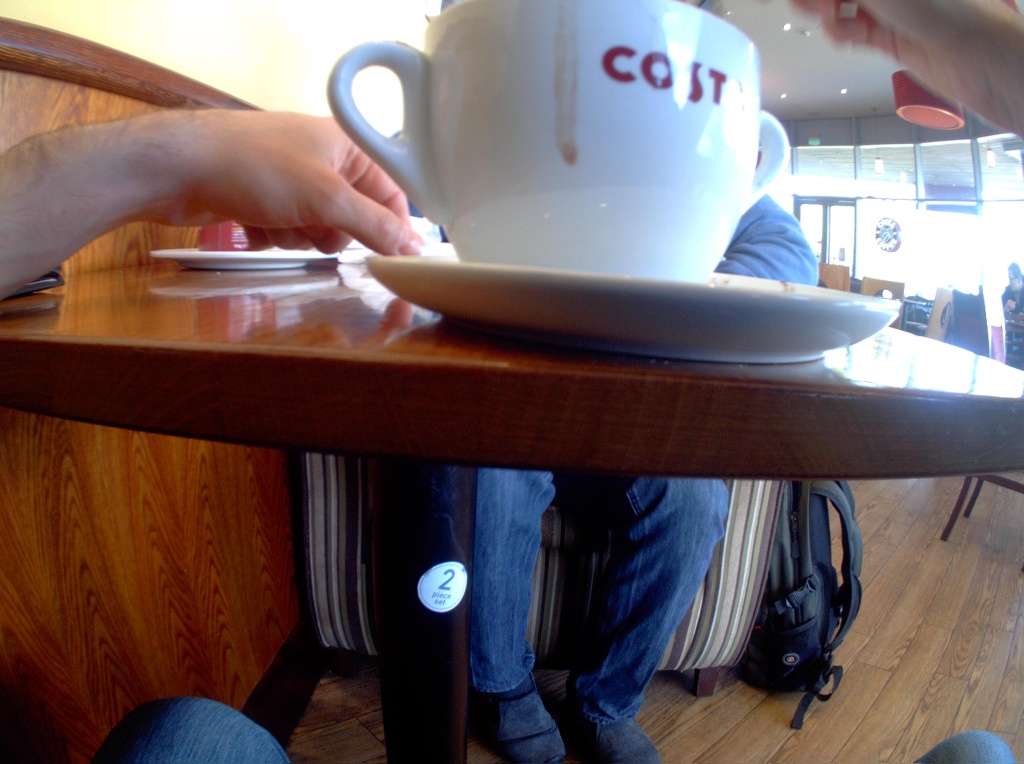}
\end{minipage}%
\begin{minipage}{\columnProportion\columnwidth}
\centering
{\helvetica Eating together}
\includegraphics[scale=\imgscale]{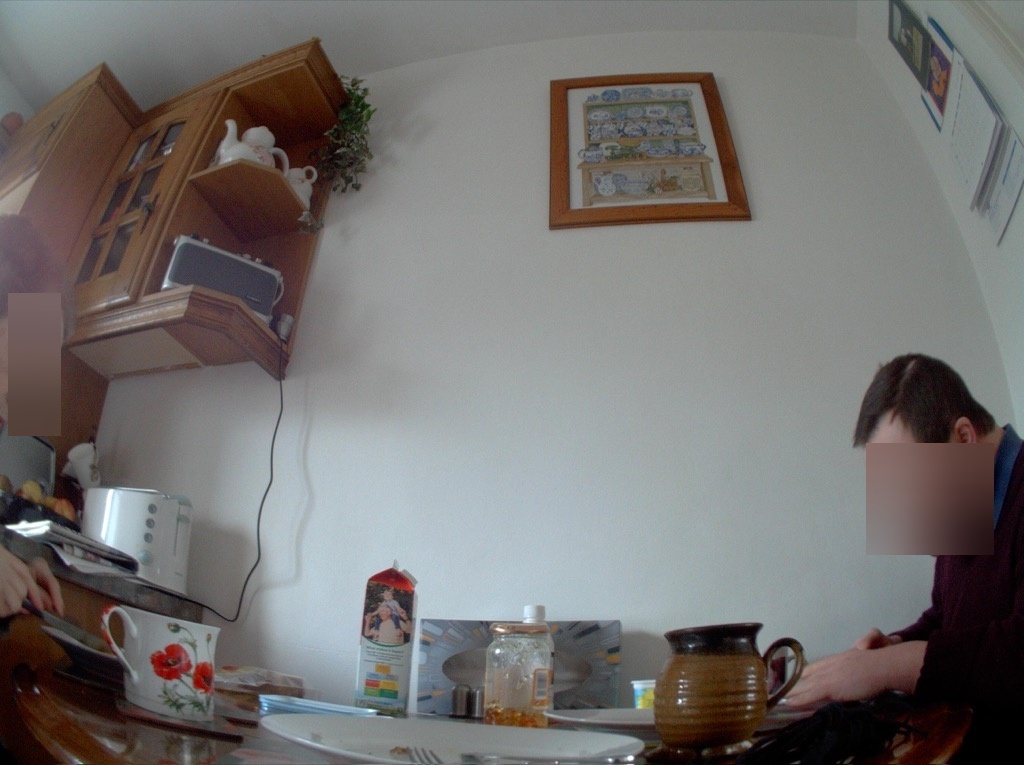}
\end{minipage}
\begin{minipage}{\columnProportion\columnwidth}
\centering
{\helvetica Walking Indoor}
\includegraphics[scale=\imgscale]{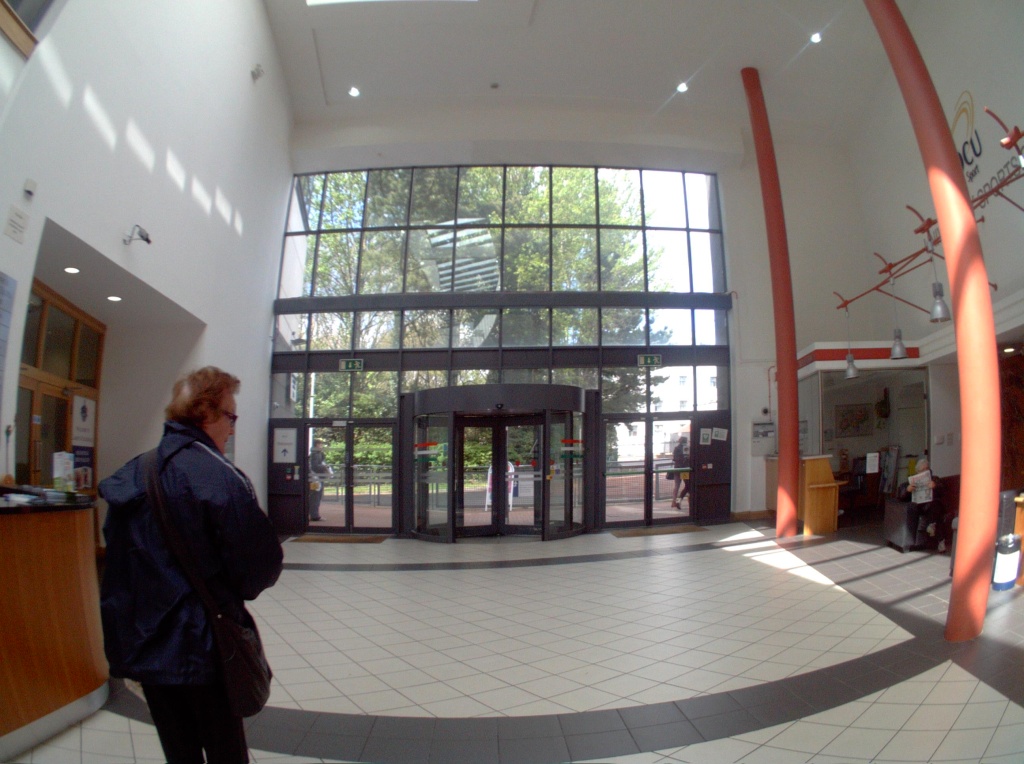}
\end{minipage}\vspace{\rowspace}%

\begin{minipage}{\columnProportion\columnwidth}
\centering
{\helvetica Mobile}
\includegraphics[scale=\imgscale]{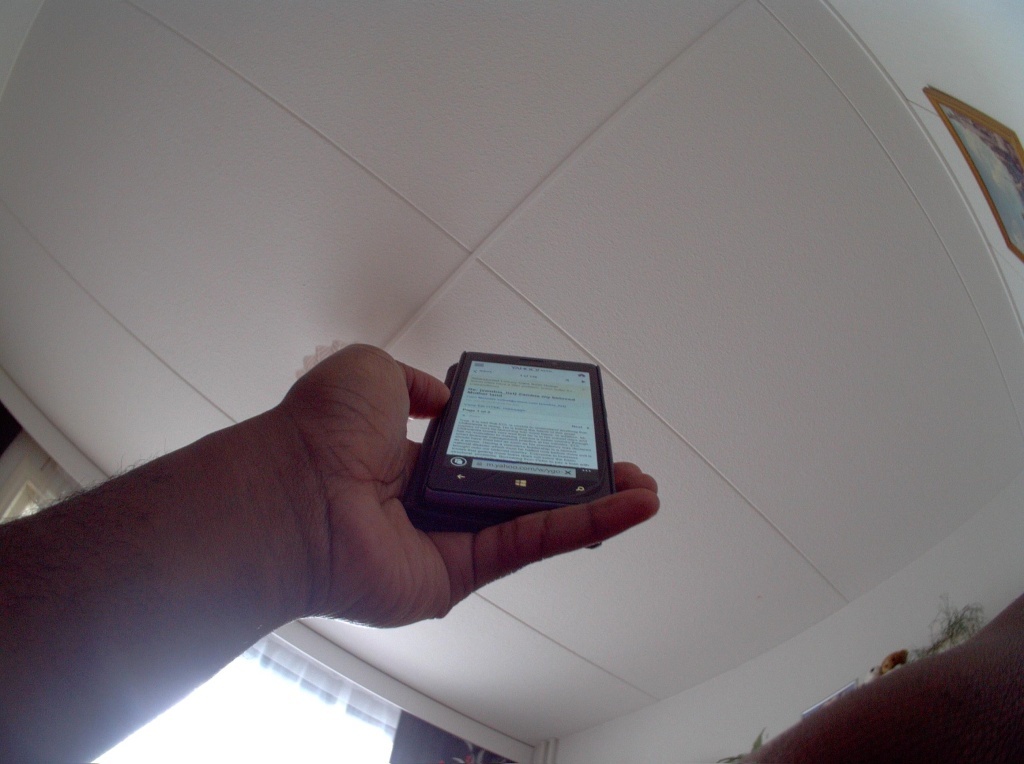}
\end{minipage}%
\begin{minipage}{\columnProportion\columnwidth}
\centering
{\helvetica Resting}
\includegraphics[scale=\imgscale]{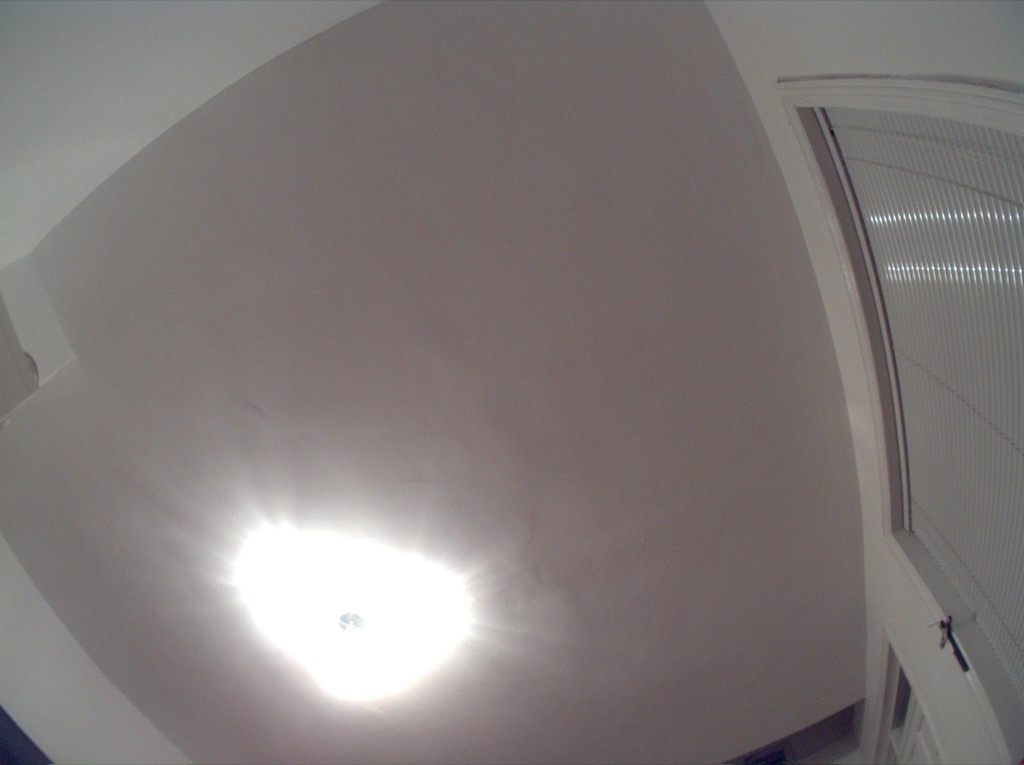}
\end{minipage}%
\begin{minipage}{\columnProportion\columnwidth}
\centering
{\helvetica Public transport}
\includegraphics[scale=\imgscale]{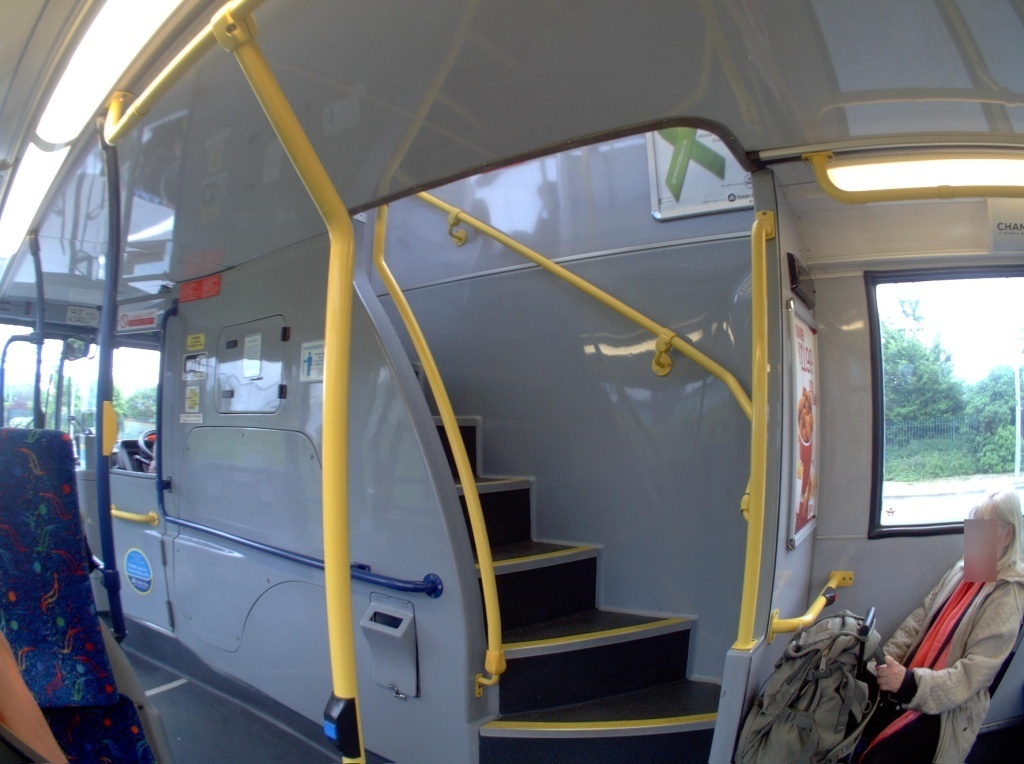}
\end{minipage}%
\begin{minipage}{\columnProportion\columnwidth}
\centering
{\helvetica Reading}
\includegraphics[scale=\imgscale]{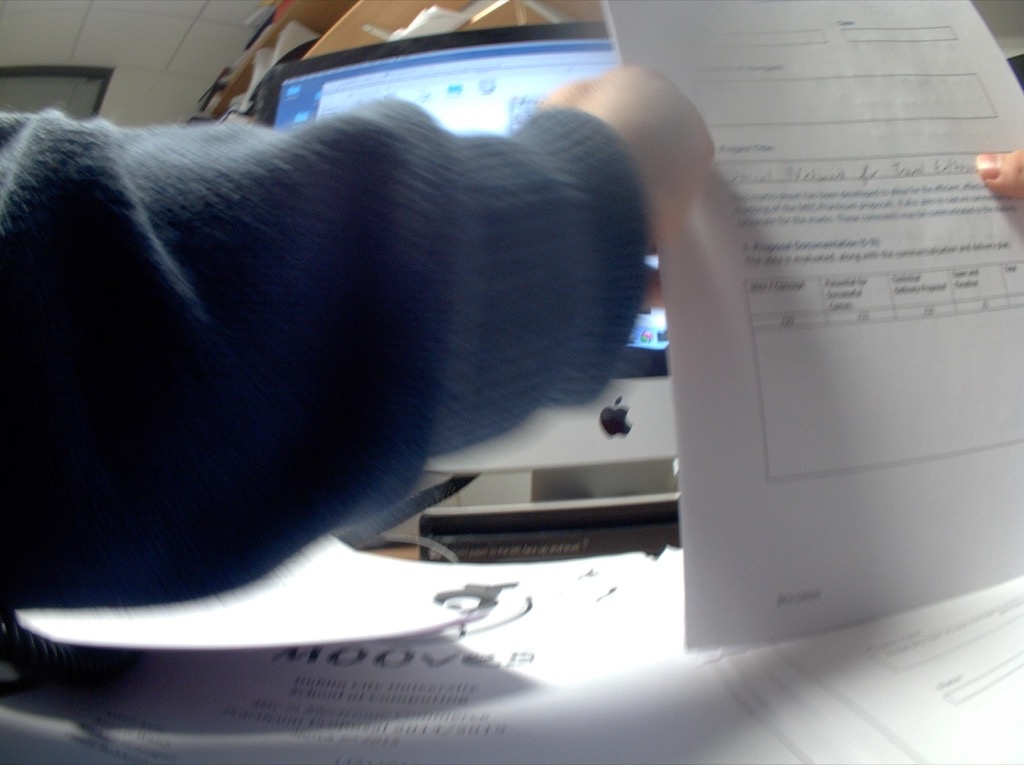}
\end{minipage}%
\begin{minipage}{\columnProportion\columnwidth}
\centering
{\helvetica Working}
\includegraphics[scale=\imgscale]{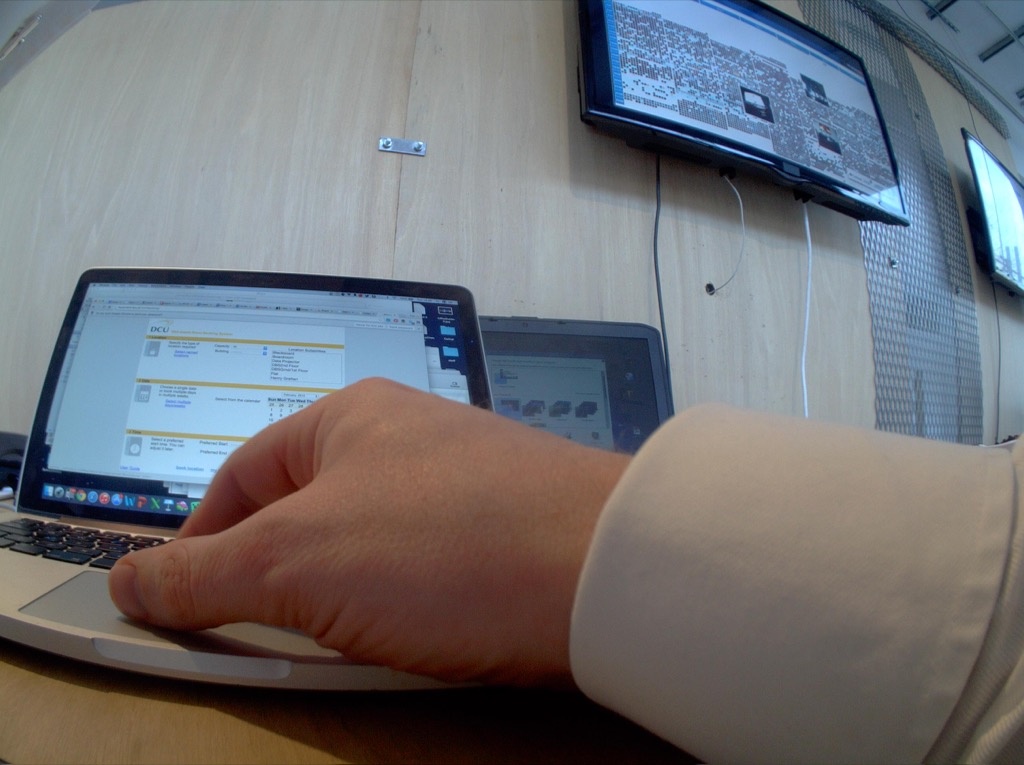}
\end{minipage}%
\begin{minipage}{\columnProportion\columnwidth}
\centering
{\helvetica Drinking/Eating Alone}
\includegraphics[scale=\imgscale]{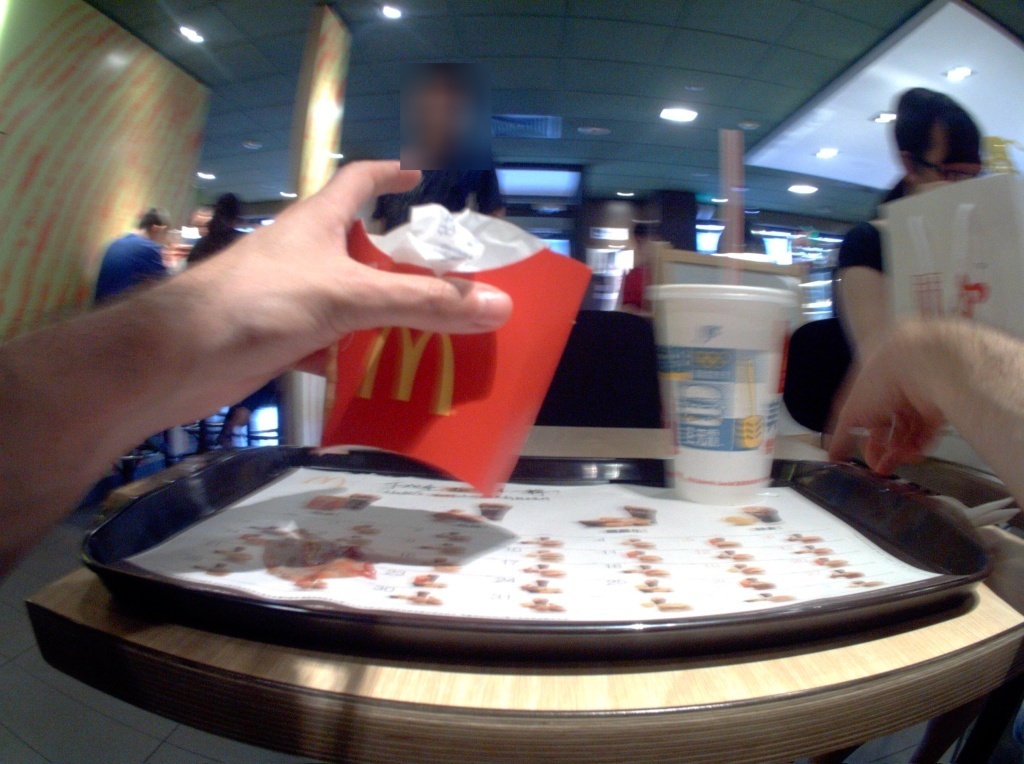}
\end{minipage}%
\begin{minipage}{\columnProportion\columnwidth}
\centering
{\helvetica Socializing}
\includegraphics[scale=\imgscale]{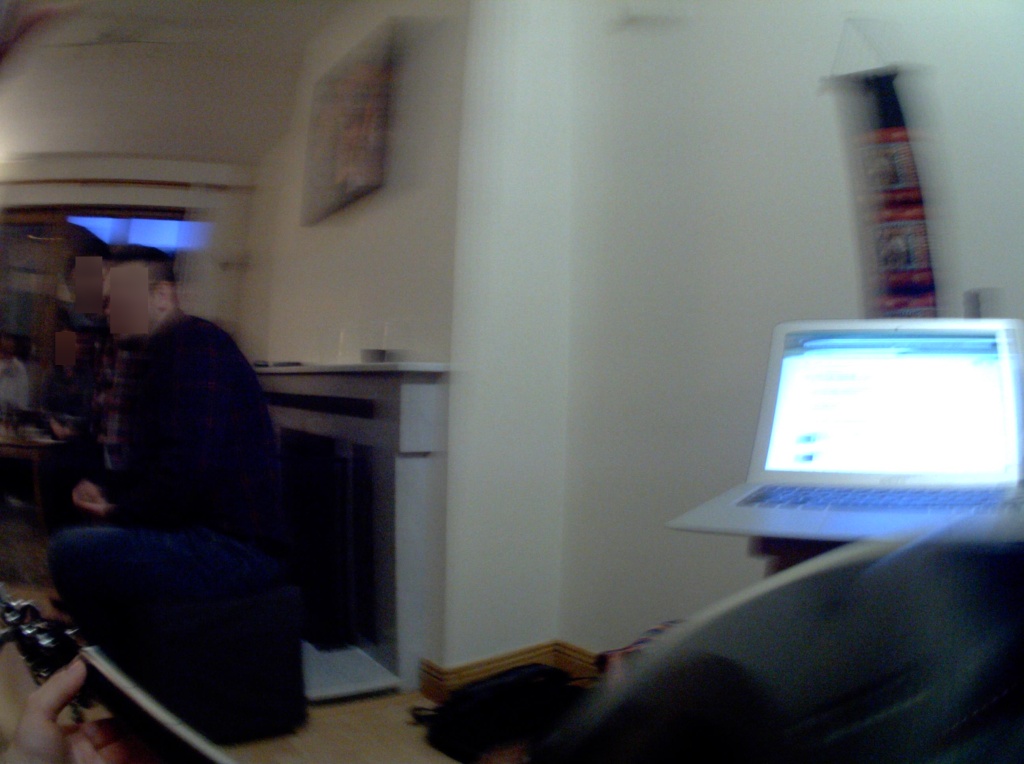}
\end{minipage}%

\end{center}
\caption[]{Examples of all activity categories from the annotated dataset.}
\label{fig:examplesActivities}
\end{center}
\end{figure*}

\begin{figure*}[!t]
\begin{center}
\includegraphics[scale=0.52]{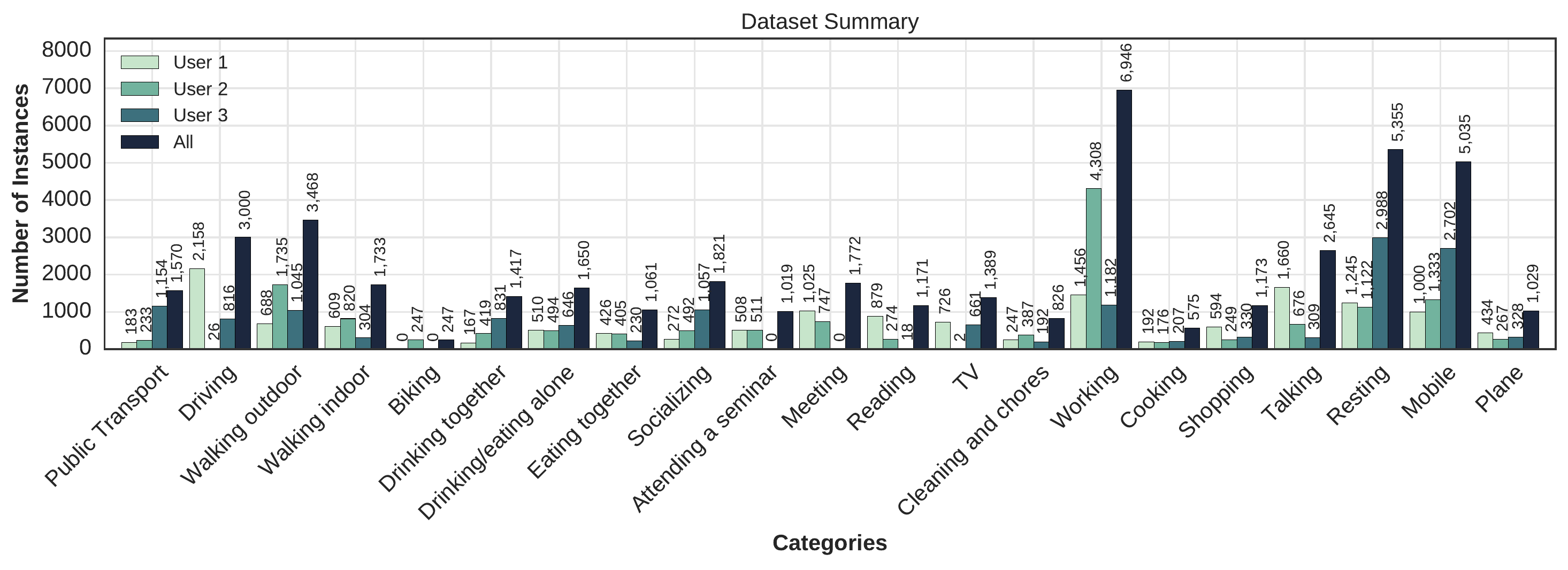}
\caption[]{Per class and per user label distribution in our dataset.}
\label{fig:overviewNtcir:datasetSummary}
\end{center}
\end{figure*}

We used a subset of 44,901 images from the NTCIR-12 dataset, that we annotated in batches with 21 activity labels, extending the dataset used in \cite{cartas2017recognizing}. These pictures correspond to all users at different dates and times, but having a similar proportion of about 15,000 images per user. We named these additional labels the UB Extended Annotations (UBEA) dataset. The complete list of activity categories and their distribution of the number of images are shown in Fig. \ref{fig:examplesActivities} and Fig. \ref{fig:overviewNtcir:datasetSummary}, respectively. The dataset splits used for the experiments of both methods are explained below.

\begin{figure*}[t]
  \centering
  \begin{subfigure}[b]{0.4\textwidth}
      \centering
      \includegraphics[scale=.8]{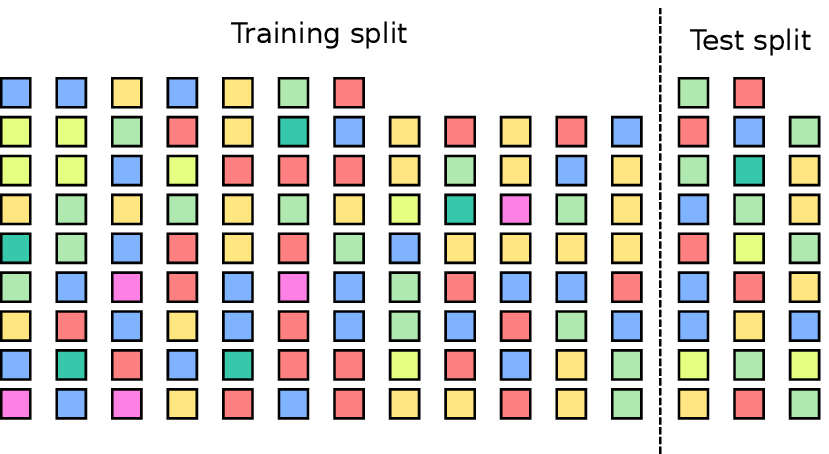}
      \caption[]{Splits used in \cite{cartas2017recognizing} for activity recognition at image level}
      \label{fig:datasetSplitExperiment1}
  \end{subfigure}\hspace{1.5cm}
  \begin{subfigure}[b]{0.4\textwidth}
      \centering
      \includegraphics[scale=0.8]{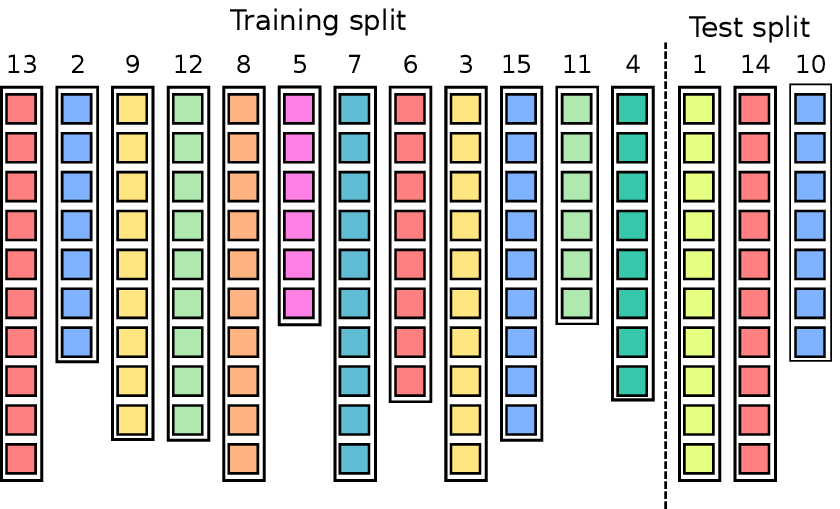}
      \caption[]{Splits used in \cite{cartas2017batch} and this work for activity recognition at batch level.}
      \label{fig:datasetSplitExperiment2}
  \end{subfigure}
\caption[]{Visual description of the dataset splits for the experiments carried out. Images of the same color belong to the same day. In (a) the split into training and test does not take into account time metadata, so that images of the same day can be found in both training and test, whereas in (b) images of the same day, can be found or in training or in test even of the split does not take into account the temporal adjacency of days so that temporally adjacent days, say 1 and 2 can be found in different sets (training or test).}
\label{fig:datasetSplit}
\end{figure*}

\textbf{Activity recognition at image level}. For this task we used the pictures from the three users for all categories regardless of their date and time as in \cite{cartas2017recognizing} and illustrated in Fig. \ref{fig:datasetSplitExperiment1} (a). In order to maintain the same percentage of samples for each class, we first performed a stratified 10-fold cross-validation over the images. Then a validation split for each fold containing 10\% of the data was created by further making a stratified shuffle of its training split.

\textbf{Activity recognition using temporal information}. For this experiment we used the same data split as in \cite{cartas2017batch}, illustrated in Fig. \ref{fig:datasetSplitExperiment2}. The purpose of this split was to separate full days of activities rather than single frames. This separation made harder the classification task since similar consecutive frames only appear in one split. Additionally, the splits proportionally maintained the class imbalance of the dataset distribution shown in Fig. \ref{fig:overviewNtcir:datasetSummary}. In order to create the splits, all the combinations of training and testing day splits were enumerated. Then, the Bhattacharyya distance between the normalized distributions of the whole dataset and each split for all combinations was calculated. Finally, the selected combination was the one with the minimal sum of Bhattacharyya distances.
 
\subsection{Evaluation metrics}
Since the dataset is highly imbalanced, the classification performance of all methods was assessed by not only using the accuracy, but also macro metrics for precision, recall, and F1-score. Since macro metrics are an unweighted average of the metrics taken separately for each class, they do not take into account of the number of instances available for each class. Moreover, the results on the activity recognition at the image level experiments are cross-validated. Considering that the photo-streams lack of event boundaries, the predictions done using temporal contextual information were calculated per frame.

\subsection{Implementation}
\label{sec:training}

\subsubsection{Activity recognition at image level}

\begin{figure}[!t] 
\begin{center} 
\includegraphics[width=0.925\columnwidth]{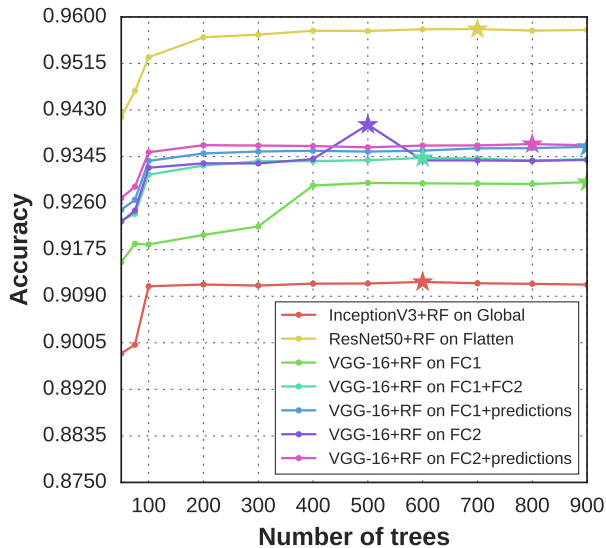} 
\caption[]{Mean accuracy on the validation set of each fold using a different number of trees. The star mark points at the maximum value. This figure is best seen in color.} 
\label{fig:acc_by_num_trees} 
\end{center} 
\end{figure}

We used three different CNN architectures as base models for our ensembles, specifically the VGG-16, InceptionV3, and ResNet 50. All these networks were pre-trained on ImageNet using the Keras framework \cite{chollet2015keras}. Moreover, we employed a class weighting scheme for all CNN models based on \cite{king_zeng_2001} to handle class imbalance during training. The random forests in our ensembles were trained using the Scikit Learning framework \cite{scikit-learn}. As input of the random forest we considered the output of each and all fully connected layers following the last convolutional layer. The training was done on the dataset split described in the previous subsection. First, we trained each CNN on all the cross-validation folds. In order to find the number of trees of each random forest, we trained all the random forests combinations of each CNN using a different number of trees. The criteria used for training them was the Gini impurity\cite{breiman1984classification} and their nodes were expanded until all the leaves were pure. All these random forest combinations were tested on each validation split and the mean accuracy for the corresponding number of trees was plotted, as shown in Fig. \ref{fig:acc_by_num_trees}. Based on this plot, the number of trees to be used was determined as the lowest one after which the performance did not improve significantly. The training details of each ensemble configuration are detailed below.

\textbf{VGG-16}. We fine-tuned a VGG-16 network~\cite{Simonyan14c} in two phases. During the first phase, only the top layers were back-propagated with the objective of initialize their weights. The optimization method used was the Stochastic Gradient Descent (SGD) for 10 epochs for all folds, a learning rate $\alpha=1\times10^{-5}$, a batch size of 1, a momentum $\mu=0.9$, and a weight decay equal to $5\times10{-6}$. In the second phase, the last three convolutional layers were also fine-tuned and the initial weights were obtained from the best epoch of the first phase. Moreover, the SGD ran for another 10 epochs for each fold and set with the same parameters except the learning rate $\alpha=4\times10^{-5}$. Additionally, dropout layers were added after each fully-connected layer as a mechanism for regularization.

\textbf{VGG-16+RF}. A set of random forests was trained using the distinct layers of the best epoch from the fine-tuned VGG-16 for each fold. The different combinations of layers for the ensembles were: FC1, FC2, FC1+FC2, FC1+softmax, and FC2+softmax. Their corresponding number of trees of each combination were 400, 500, 600, 300, and 200. The maximum depth of each combination was 49, 47, 53, 58, and 48, respectively.

\begin{table*}[!h]
\caption{Comparison of the ensembles of CNN+Random forest on different combinations of layers. Upper table shows the recall per class and the lower table shows the performance metrics.}
\label{tab:classificationComparison}
\centering
\resizebox{2.05\columnwidth}{!}{%
\begin{tabular}{ | l | c |c |c |c |c |c |c |c |c |c |}
\hline
\multirow{2}{*}{\textbf{Activity}} & \multirow{2}{*}{\textbf{InceptionV3}} & \textbf{InceptionV3+} & \multirow{2}{*}{\textbf{ResNet50}} & \textbf{ResNet50+} & \multirow{2}{*}{\textbf{VGG-16}} & \textbf{VGG-16+} & \textbf{VGG-16+RF} & \textbf{VGG-16+RF} & \textbf{VGG-16+} & \textbf{VGG-16+RF} \\
 &  & \textbf{RF on GAP} &  & \textbf{RF on AP} &  & \textbf{RF on FC1} & \textbf{on FC1+FC2} & \textbf{on FC1+Pred} & \textbf{RF on FC2} & \textbf{on FC2+Pred} \\ \hline
Public Transport & 89.23 & 91.46 &  \textbf{92.41}  & 91.59 & 89.17 & 90.87 & 91.14 & 91.53 & 91.27 & 91.59 \\ \hline Driving & 99.60 & 99.83 &  \textbf{99.90}  & 99.83 & 99.50 & 99.74 & 99.77 & 99.77 & 99.73 & 99.77 \\ \hline Walking outdoor & 90.95 & 97.35 & 94.61 &  \textbf{98.27}  & 90.89 & 96.96 & 97.03 & 96.94 & 97.00 & 96.83 \\ \hline Walking indoor & 79.17 & 95.15 & 92.73 &  \textbf{95.85}  & 79.63 & 94.81 & 94.92 & 94.75 & 95.04 & 94.87 \\ \hline Biking & 91.93 & 98.80 & 99.17 &  \textbf{99.20}  & 92.32 & 98.20 & 97.57 & 97.98 & 97.17 & 97.98 \\ \hline Drinking together & 80.74 & 95.63 & 94.93 & 95.49 & 89.42 &  \textbf{96.48}  & 95.98 & 96.05 & 95.91 & 95.84 \\ \hline Drinking/eating alone & 75.94 &  \textbf{90.85}  & 88.25 & 90.37 & 74.92 & 88.70 & 89.77 & 89.71 & 89.65 & 89.71 \\ \hline Eating together & 80.58 & 96.50 & 96.42 &  \textbf{98.11}  & 88.78 & 96.75 & 97.26 & 97.26 & 96.89 & 97.36 \\ \hline Socializing & 89.13 & 97.37 & 97.47 & 98.63 & 91.60 & 98.47 & 98.46 & 98.46 & 98.57 &  \textbf{98.73}  \\ \hline Attending a seminar & 78.70 & 78.70 &  \textbf{80.57}  & 78.31 & 76.93 & 77.75 & 78.21 & 78.11 & 78.21 & 78.11 \\ \hline Meeting & 80.92 & 93.29 & 96.61 &  \textbf{96.78}  & 90.00 & 95.36 & 95.49 & 95.49 & 95.54 & 95.43 \\ \hline Reading & 81.40 & 97.35 & 96.83 &  \textbf{97.69}  & 88.23 & 96.49 & 96.67 & 96.67 & 96.59 & 96.59 \\ \hline TV & 93.74 & 96.55 & 96.04 & 96.54 & 92.59 &  \textbf{97.21}  & 96.91 & 96.69 & 97.05 & 96.91 \\ \hline Cleaning and chores & 57.14 & 92.61 & 92.22 & 92.39 & 61.84 & 92.06 & 92.13 & 92.62 & 92.26 &  \textbf{92.74}  \\ \hline Working & 95.82 & 98.81 & 97.34 &  \textbf{99.38}  & 96.27 & 98.18 & 97.98 & 97.91 & 97.96 & 97.91 \\ \hline Cooking & 81.98 & 95.10 & 91.69 & 96.15 & 84.78 & 96.48 & 95.97 & 95.79 & 95.47 &  \textbf{96.49}  \\ \hline Shopping & 81.07 & 95.40 & 93.18 &  \textbf{96.68}  & 82.44 & 94.98 & 95.57 & 95.14 & 95.57 & 95.65 \\ \hline Talking & 72.36 &  \textbf{95.54}  & 89.64 & 94.10 & 76.25 & 93.53 & 93.50 & 93.50 & 93.65 & 93.61 \\ \hline Resting & 87.97 & 91.07 & 91.54 & 91.02 & 92.87 & 94.56 & 94.90 &  \textbf{95.14}  & 94.90 & 95.11 \\ \hline Mobile & 89.59 & 97.22 & 95.27 &  \textbf{97.58}  & 93.32 & 97.55 & 97.46 & 97.36 & 97.30 & 97.34 \\ \hline Plane &  \textbf{87.67}  & 87.16 & 82.61 & 75.70 & 84.94 & 77.21 & 80.66 & 81.15 & 81.54 & 83.58 \\ \hline  \noalign{\vskip 0.25cm} \hline
 \textbf{Accuracy} &  87.07   &  95.27   &  94.08   &  95.39   &  89.46   &  95.26   &  95.41   &  95.41   &  95.41   &  \textbf{95.50}   \\ \hline  \textbf{Macro precision} &  83.85   &  94.96   &  92.60   &  \textbf{96.14}   &  87.06   &  94.63   &  94.59   &  94.62   &  94.57   &  94.61   \\ \hline  \textbf{Macro recall} &  84.08   &  94.37   &  93.31   &  94.27   &  86.51   &  93.92   &  94.16   &  94.19   &  94.15   &  \textbf{94.39}   \\ \hline  \textbf{Macro F1-score} &  83.77   &  94.53   &  92.78   &  \textbf{95.00}   &  86.45   &  94.11   &  94.23   &  94.28   &  94.23   &  94.38   \\ \hline 
\end{tabular}
}
\end{table*}

\textbf{InceptionV3}. InceptionV3 was also fine-tuned in two phases. During the first phase, the last fully-connected layer was optimized using SGD for 10 epochs for all folds, a learning rate $\alpha=1\times10^{-5}$, a batch size of 32, a momentum $\mu=0.9$, and a weight decay equal to $5\times10{-6}$. During the second phase, the last inception block was added to the optimization process and the network was optimized for another 10 epochs setting the learning rate $\alpha=4\times10^{-5}$ and the batch size to 10. Moreover, dropout layers were added before and after the last global average pooling layer.

\textbf{InceptionV3+RF}. A random forest was trained using the global average pooling layer from InceptionV3 pre-trained on ImageNet. This random forest had 100 trees as estimators and a maximum depth of 58.

\textbf{ResNet50}. ResNet50 was fine-tuned in two phases. The last fully-connected layer was optimized in the first phase using SGD for 10 epochs for all folds, a learning rate $\alpha=1\times10^{-3}$, a batch size of 32, a momentum $\mu=0.9$, and a weight decay equal to $5\times10{-6}$. During the last phase, the last residual block was also optimized using SGD with same learning rate and a batch size of 10 for three additional epochs.

\textbf{ResNet50+RF}. A random forest was trained using the last activation layer from an InceptionV3 network pre-trained on ImageNet. This random forests had 400 trees as estimators and a maximum depth of 49.

\begin{table}[!ht]
\begin{center}
\resizebox{1.0\columnwidth}{0.4446\textheight}{%
\begin{tabular}{ | c | c |c |c |c |}
\hline
\multirow{2}{*}{Method} & \multirow{2}{*}{Accuracy} & Macro & Macro & Macro \\
& & Precision & Recall & F1-score \\ \hline
ResNet50& 78.44 & 72.44 & 70.62 & 69.85 \\ \hline
ResNet50+RF& \multirow{2}{*}{74.62} & \multirow{2}{*}{76.64} & \multirow{2}{*}{62.61} & \multirow{2}{*}{64.94} \\
on GAP& & & & \\ \hline
ResNet50+RF on& \multirow{2}{*}{73.09} & \multirow{2}{*}{71.97} & \multirow{2}{*}{59.52} & \multirow{2}{*}{62.51} \\
GAP timestep 5& & & & \\ \hline
ResNet50+RF on& \multirow{2}{*}{71.29} & \multirow{2}{*}{72.12} & \multirow{2}{*}{57.23} & \multirow{2}{*}{60.16} \\
GAP timestep 10& & & & \\ \hline
ResNet50+RF+LSTM& \multirow{2}{*}{81.10} & \multirow{2}{*}{75.69} & \multirow{2}{*}{72.90} & \multirow{2}{*}{72.41} \\
timestep 5& & & & \\ \hline
ResNet50+RF+LSTM& \multirow{2}{*}{85.12} & \multirow{2}{*}{81.48} & \multirow{2}{*}{78.98} & \multirow{2}{*}{78.76} \\
timestep 10& & & & \\ \hline
ResNet50+RF+LSTM& \multirow{2}{*}{83.29} & \multirow{2}{*}{80.75} & \multirow{2}{*}{78.04} & \multirow{2}{*}{77.24} \\
timestep 15& & & & \\ \hline \hline
InceptionV3& 78.31 & 73.06 & 69.94 & 70.60 \\ \hline
InceptionV3+RF& \multirow{2}{*}{77.08} & \multirow{2}{*}{73.68} & \multirow{2}{*}{67.24} & \multirow{2}{*}{68.54} \\
on AP& & & & \\ \hline
InceptionV3+RF& \multirow{2}{*}{78.82} & \multirow{2}{*}{75.62} & \multirow{2}{*}{69.11} & \multirow{2}{*}{69.93} \\
timestep 5& & & & \\ \hline
InceptionV3+RF& \multirow{2}{*}{78.55} & \multirow{2}{*}{75.93} & \multirow{2}{*}{69.12} & \multirow{2}{*}{69.89} \\
timestep 10& & & & \\ \hline
InceptionV3+RF+LSTM& \multirow{2}{*}{86.04} & \multirow{2}{*}{85.58} & \multirow{2}{*}{83.74} & \multirow{2}{*}{83.56} \\
on AP timestep 5& & & & \\ \hline
\textbf{InceptionV3+RF+LSTM}& \multirow{2}{*}{\textbf{89.85}} & \multirow{2}{*}{\textbf{89.58}} & \multirow{2}{*}{\textbf{87.20}} & \multirow{2}{*}{\textbf{87.22}} \\
\textbf{on AP timestep 10}& & & & \\ \hline
InceptionV3+RF+LSTM& \multirow{2}{*}{87.72} & \multirow{2}{*}{86.95} & \multirow{2}{*}{84.44} & \multirow{2}{*}{84.37} \\
on AP timestep 15& & & & \\ \hline \hline
VGG-16 & 75.97 & 68.50 & 67.49 & 66.80 \\ \hline
VGG-16+RF& \multirow{2}{*}{74.22} & \multirow{2}{*}{70.77} & \multirow{2}{*}{65.06} & \multirow{2}{*}{65.80} \\
on FC1& & & & \\ \hline
VGG-16+RF on& \multirow{2}{*}{76.80} & \multirow{2}{*}{71.93} & \multirow{2}{*}{66.86} & \multirow{2}{*}{67.51} \\
FC1+Pred\cite{cartas2017recognizing} & & & & \\ \hline
VGG-16+RF& \multirow{2}{*}{75.02} & \multirow{2}{*}{69.12} & \multirow{2}{*}{65.60} & \multirow{2}{*}{65.60} \\
on FC2& & & & \\ \hline
VGG-16+RF on& \multirow{2}{*}{75.87} & \multirow{2}{*}{69.78} & \multirow{2}{*}{66.08} & \multirow{2}{*}{66.28} \\
FC2+Pred& & & & \\ \hline
VGG-16+RF on Softmax +& \multirow{2}{*}{77.39} & \multirow{2}{*}{69.66} & \multirow{2}{*}{67.79} & \multirow{2}{*}{66.99} \\
Date\&Time + Color \cite{castro2015predicting} & & & & \\ \hline
VGG-16+RF on FC1 +& 76.91 & 72.07 & 66.77 & 67.27 \\
Softmax + Date\&Time & & & & \\ \hline%
VGG-16+RF on& \multirow{2}{*}{76.04} & \multirow{2}{*}{74.44} & \multirow{2}{*}{65.98} & \multirow{2}{*}{66.93} \\
FC1 timestep 5& & & & \\ \hline
VGG-16+RF on& \multirow{2}{*}{75.67} & \multirow{2}{*}{75.20} & \multirow{2}{*}{64.93} & \multirow{2}{*}{66.26} \\
FC1 timestep 10& & & & \\ \hline%
VGG-16+RF on& \multirow{2}{*}{77.43} & \multirow{2}{*}{74.54} & \multirow{2}{*}{67.13} & \multirow{2}{*}{67.57} \\
FC1+Pred timestep 5& & & & \\ \hline
VGG-16+RF on& \multirow{2}{*}{76.66} & \multirow{2}{*}{74.10} & \multirow{2}{*}{65.91} & \multirow{2}{*}{66.67} \\
FC1+Pred timestep 10& & & & \\ \hline%
VGG-16+RF on& \multirow{2}{*}{76.83} & \multirow{2}{*}{73.48} & \multirow{2}{*}{67.92} & \multirow{2}{*}{67.92} \\
FC2 timestep 5& & & & \\ \hline
VGG-16+RF on& \multirow{2}{*}{76.49} & \multirow{2}{*}{73.40} & \multirow{2}{*}{67.72} & \multirow{2}{*}{67.61} \\
FC2 timestep 10& & & & \\ \hline%
VGG-16+RF on& \multirow{2}{*}{77.79} & \multirow{2}{*}{74.60} & \multirow{2}{*}{68.31} & \multirow{2}{*}{68.53} \\
FC2+Pred timestep 5& & & & \\ \hline
VGG-16+RF on& \multirow{2}{*}{77.75} & \multirow{2}{*}{74.82} & \multirow{2}{*}{67.89} & \multirow{2}{*}{68.37} \\
FC2+Pred timestep 10& & & & \\ \hline
VGG-16+LSTM \cite{cartas2017batch} & \multirow{2}{*}{79.68} & \multirow{2}{*}{72.96} & \multirow{2}{*}{71.36} & \multirow{2}{*}{70.87} \\
on FC1 timestep 5 & & & & \\ \hline
VGG-16+LSTM \cite{cartas2017batch} & \multirow{2}{*}{80.39} & \multirow{2}{*}{75.25} & \multirow{2}{*}{71.86} & \multirow{2}{*}{71.97} \\
timestep 10 & & & & \\ \hline
VGG-16+LSTM \cite{cartas2017batch} & \multirow{2}{*}{81.73} & \multirow{2}{*}{76.68} & \multirow{2}{*}{74.04} & \multirow{2}{*}{74.16} \\
timestep 15 & & & & \\ \hline%
VGG-16+RF+LSTM& \multirow{2}{*}{85.96} & \multirow{2}{*}{79.81} & \multirow{2}{*}{81.36} & \multirow{2}{*}{80.00} \\
FC1 timestep 5& & & & \\ \hline
VGG-16+RF+LSTM& \multirow{2}{*}{86.87} & \multirow{2}{*}{80.45} & \multirow{2}{*}{81.36} & \multirow{2}{*}{80.39} \\
FC1 timestep 10& & & & \\ \hline
VGG-16+RF+LSTM& \multirow{2}{*}{85.55} & \multirow{2}{*}{80.00} & \multirow{2}{*}{79.34} & \multirow{2}{*}{78.45} \\
FC1 timestep 15& & & & \\ \hline
VGG-16+RF+LSTM& \multirow{2}{*}{85.45} & \multirow{2}{*}{79.29} & \multirow{2}{*}{79.11} & \multirow{2}{*}{78.05} \\
FC1+Pred timestep 5& & & & \\ \hline
VGG-16+RF+LSTM& \multirow{2}{*}{87.71} & \multirow{2}{*}{81.55} & \multirow{2}{*}{81.88} & \multirow{2}{*}{81.10} \\
FC1+Pred timestep 10& & & & \\ \hline
VGG-16+RF+LSTM& \multirow{2}{*}{86.24} & \multirow{2}{*}{80.05} & \multirow{2}{*}{80.79} & \multirow{2}{*}{79.56} \\
FC1+Pred timestep 15& & & & \\ \hline%
VGG-16+RF+LSTM& \multirow{2}{*}{85.73} & \multirow{2}{*}{81.22} & \multirow{2}{*}{80.56} & \multirow{2}{*}{79.96} \\
FC2 timestep 5& & & & \\ \hline
VGG-16+RF+LSTM& \multirow{2}{*}{86.78} & \multirow{2}{*}{81.37} & \multirow{2}{*}{80.97} & \multirow{2}{*}{80.37} \\
FC2 timestep 10& & & & \\ \hline
VGG-16+RF+LSTM& \multirow{2}{*}{85.38} & \multirow{2}{*}{79.71} & \multirow{2}{*}{79.80} & \multirow{2}{*}{78.63} \\
FC2 timestep 15& & & & \\ \hline
VGG-16+RF+LSTM& \multirow{2}{*}{85.89} & \multirow{2}{*}{79.62} & \multirow{2}{*}{80.35} & \multirow{2}{*}{78.82} \\
FC2+Pred timestep 5& & & & \\ \hline
VGG-16+RF+LSTM& \multirow{2}{*}{86.85} & \multirow{2}{*}{80.22} & \multirow{2}{*}{81.19} & \multirow{2}{*}{80.23} \\
FC2+Pred timestep 10& & & & \\ \hline
VGG-16+RF+LSTM& \multirow{2}{*}{86.35} & \multirow{2}{*}{80.11} & \multirow{2}{*}{80.74} & \multirow{2}{*}{79.51} \\
FC2+Pred timestep 15& & & & \\ \hline
\end{tabular}
}
\end{center}
\caption{Performance summary of the experiments on activity recognition using temporal contextual information.}
\label{tab:performanceSummaryExperiment2}
\end{table}
\subsubsection{Activity recognition using temporal information}

\textbf{VGG-16}. The VGG-16 was trained for 14 epochs using the SGD algorithm. During the first 10 epochs only the fully connected layers were optimized using a learning rate $\alpha=1\times10^{-5}$, a batch size of 1, a momentum $\mu=0.9$, and a weight decay equal to $5\times10{-6}$. During the last 4 epochs, the last 3 convolutional layers were also fine-tuned and the learning rate changed to $\alpha=1\times10^{-5}$. In comparison with the previous experiment, no class weighting scheme was used for training.

\textbf{VGG-16+RF}. Five different combinations of random forests using 500 estimators were trained for this experiment. The first four models were previous combinations of layers from the first experiment, specifically, FC1, FC2, FC1+Prediction, and FC2+Prediction. The max. depth of each combination was 55, 52, 60, and 51, correspondingly. Another model was trained for comparison purposes following the description in \cite{castro2015predicting}. In other words, its input was a feature vector obtained by concatenating the softmax probability scores, the day of the week, the time of the day, and 10-bin size histogram for each color channel. Its resulting max. depth was 47. The last trained model used the FC1 layer, the softmax probability scores, the day of the week, the hour, and the time. Its maximum depth was 57.

\begin{figure*}[!t]
\begin{center}
\begin{minipage}[b]{.76\textwidth}
\includegraphics[trim={0.4cm 8cm 0.3cm 0.8cm},clip,scale=0.6]{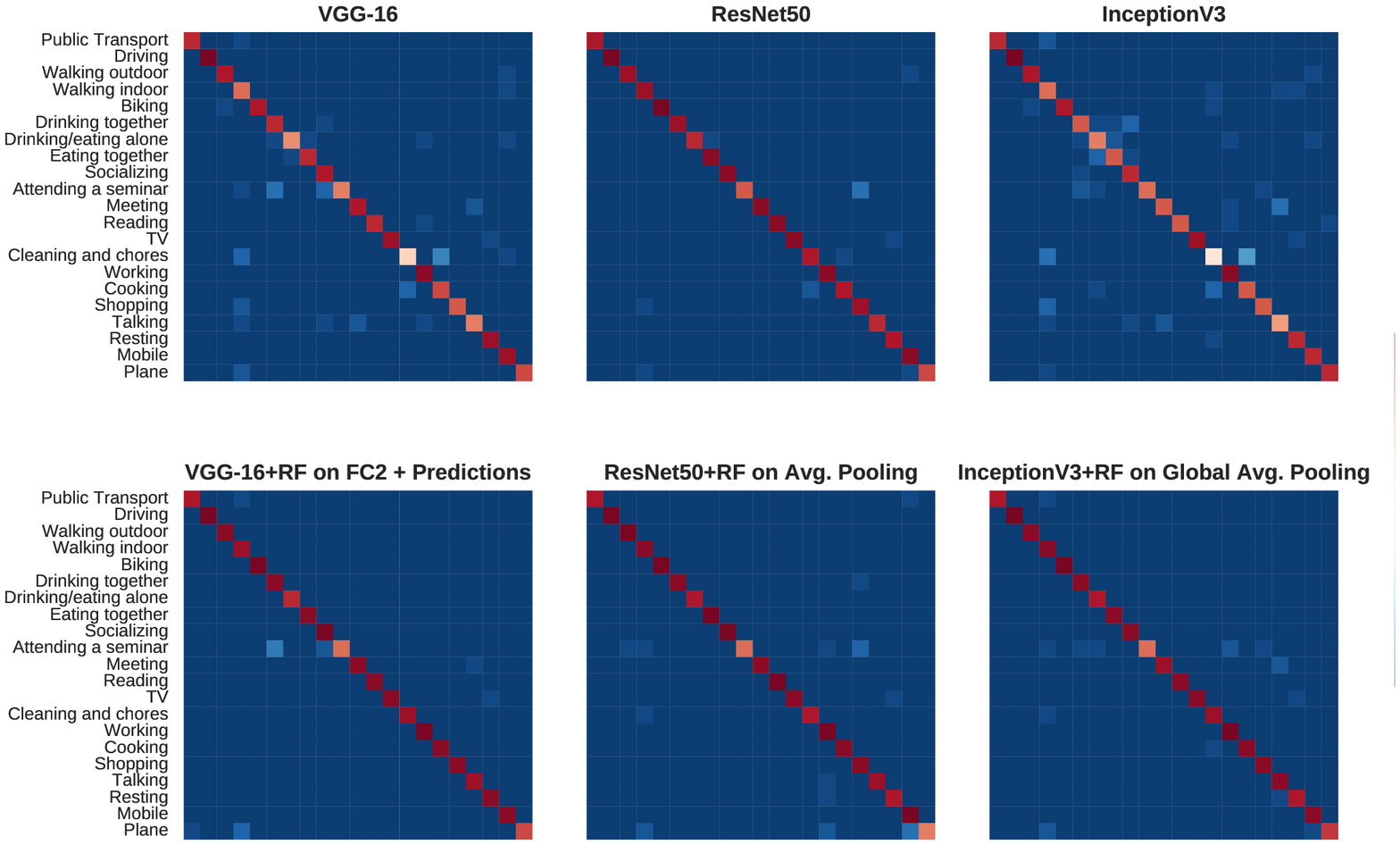}

\includegraphics[trim={0.4cm 0.8cm 0.3cm 8cm},clip,scale=0.6]{confusion_matrices.eps}
\end{minipage}
\begin{minipage}[b]{.05\textwidth}
\includegraphics[scale=0.6]{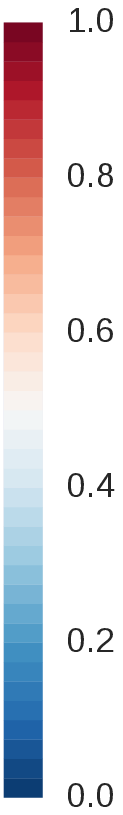}
\end{minipage}
\caption[]{Normalized confusion matrices of the best combination of layers for each baseline convolutional neural network. This figure is best seen in color.}
\label{fig:confusion_matrices}
\end{center}
\end{figure*}

\textbf{VGG-16+RF+LSTM}. For this configuration we trained a LSTM using the predictions from the ensemble of VGG-16 plus the RF on FC1. The LSTM had 32 units and its output was connected to a fully-connected layer. Dropout layers were added between the input and output of the LSTM layer as a way to perform regularization. We trained three configurations using batches of consecutive frames, obtained by using a temporal sliding window of size (or \textit{timestep}) 5, 10, and 15.

\textbf{VGG-16+RF many-to-one}. Four different random forests were trained by concatenating the output vectors of the FC1, FC2, and predictions of the VGG-16 network of a consecutive number of frames. This configuration employed the same sampling window strategy for training. For example, Fig. \ref{fig:temporal_many2one_RF} shows a 5 timestep configuration training on two consecutive batches and predicting the activity from the last frame. On this architecture, the same label is assigned to all frames considered. We considered configurations with timestep 5 and 10. The number of estimators for FC1, FC1 + Prediction, FC2, and FC2 + Prediction were 700. The resulting maximum depth of each combination for timestep 5 was 57, 56, 54, and 53; and for timestep 10 was 53, 60, 53, and 53.

\textbf{InceptionV3}. This network was trained for 12 epochs using the SGD algorithm. During the first 10 epochs only the last fully-connected layer was optimized using a learning rate $\alpha=1\times10^{-5}$, a batch size of 10, a momentum $\mu=0.9$, and a weight decay equal to $5\times10{-6}$. During the last 2 epochs, the last inception block was also fine-tuned and the learning rate changed to $\alpha=4\times10^{-5}$. No class weighting scheme was used for training.

\textbf{InceptionV3+RF}. The global average pooling (GAP) layer was used as the input of a random forests using 100 estimators and its maximum depth was 72.

\textbf{InceptionV3+RF+LSTM}. We trained a LSTM using the predictions from the ensemble of InceptionV3 plus the RF on GAP layer. The LSTM had 32 units and its output was connected to a fully-connected layer. Dropout layers were added between the input and output of the LSTM layer as a way to perform regularization. We trained three configurations using batches of consecutive frames, obtained by using a temporal sliding window of \textit{timestep} 5, 10, and 15.
	
\textbf{InceptionV3+RF many-to-one}. One random forest was trained by joining the output vector of the GAP layer of the InceptionV3 network of a consecutive number of frames. This configuration employed the same sampling window strategy for training. We considered configurations with timestep 5 and 10. The number of estimators for both cases was 700. The maximum depth obtained for both timesteps was 62 and 66, correspondingly.

\textbf{ResNet50}. This network was trained for 4 epochs using the SGD algorithm. The last fully-connected layer was optimized in the first phase using SGD for 2 epochs for all folds, a learning rate $\alpha=1\times10^{-3}$, a batch size of 10, a momentum $\mu=0.9$, and a weight decay equal to $5\times10{-6}$. During the last 2 epochs, the last inception block was also fine-tuned using the same values. No class weighting scheme was used for training.

\textbf{ResNet50+RF}. A random forest was trained on the AP layer using 500 estimators and its resulting max. depth was 53.

\textbf{ResNet50+RF+LSTM}. We trained a LSTM using the predictions from the trained ensemble. The LSTM had 32 units and its output was connected to a fully-connected layer. Dropout layers were added between the input and output of the LSTM layer as a way to perform regularization. We trained three configurations using batches of consecutive frames, obtained by using \textit{timesteps} of 5, 10, and 15.

\textbf{ResNet50+RF many-to-one}. One random forest was trained by concatenating the output vector of the AP layer of the ResNet50 network of a consecutive number of frames. This configuration employed the same sampling window strategy for training. We considered configurations with timestep 5 and 10. The number of estimators for both cases was 700. Moreover, the maximum depth for the configurations with timestep 5 and 10 were 65 and 63, respectively.

\subsection{Results and discussion}
\label{sec:results}

\begin{figure*}[!t]
\newcommand\imgscale{0.085}
\newcommand\spaceBetweenBoxes{0.3cm}
\newcommand\columnProportion{0.9}
\newcommand\boxProportion{1}
\newcommand\columnBox{1cm}
\newcommand\spaceBetweenColumns{0.5cm}

\centering

\begin{minipage}{.2\textwidth}
\centering
\begin{minipage}{\columnProportion\textwidth}
\centering
{\helvetica Reading}
\includegraphics[scale=\imgscale]{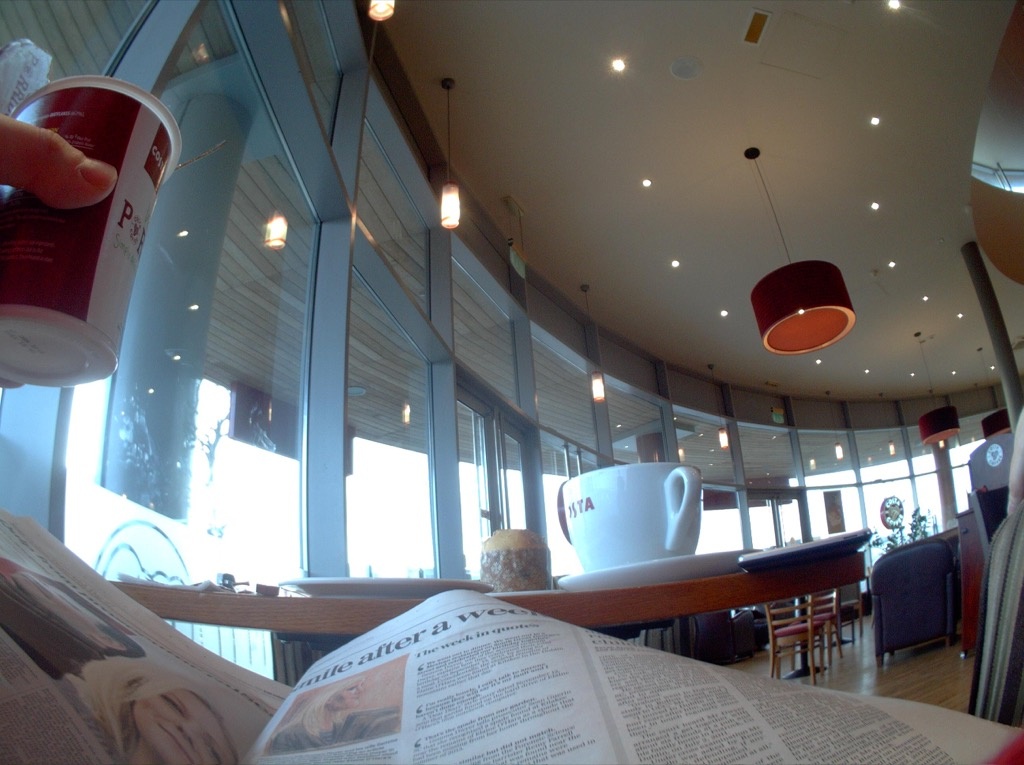}
\end{minipage}\vspace{\spaceBetweenBoxes}

\begin{minipage}{\columnProportion\textwidth}
\centering
{\helvetica VGG-16 Top 5}
\resizebox{\boxProportion\columnwidth}{!}{%
\begin{tabular}{|c|c|c|}
\hline
\textbf{\#} & \textbf{Activity} & \textbf{Score} \\ \hline
\textbf{\#} & \textbf{Activity} & \textbf{Score} \\ \hline
1 & Drinking/eating alone & 0.9853\\ \hline
\cellcolor{green!25}2 & \cellcolor{green!25}Reading & \cellcolor{green!25}0.0100\\ \hline
3 & Drinking together & 0.0033\\ \hline
4 & Eating together & 0.0012\\ \hline
5 & Mobile & 0.00006\\ \hline
\end{tabular}
}
\end{minipage}\vspace{\spaceBetweenBoxes}

\begin{minipage}{\columnProportion\textwidth}
\centering
{\helvetica ResNet Top 5}
\resizebox{\boxProportion\columnwidth}{!}{%
\begin{tabular}{|c|c|c|}
\hline
\textbf{\#} & \textbf{Activity}  & \textbf{Score} \\ \hline
1& Drinking/eating alone& 0.9888\\ \hline
\cellcolor{green!25}2& \cellcolor{green!25}Reading& \cellcolor{green!25}0.0112\\ \hline
3& Socializing& 0.000\\ \hline
4& Public Transport& 0.0000\\ \hline
5& Eating together& 0.0000\\ \hline
\end{tabular}
}
\end{minipage}\vspace{\spaceBetweenBoxes}

\begin{minipage}{\columnProportion\textwidth}
\centering
{\helvetica InceptionV3 Top 5}
\resizebox{\boxProportion\columnwidth}{!}{%
\begin{tabular}{|c|c|c|}
\hline
\textbf{\#} & \textbf{Activity}  & \textbf{Score} \\ \hline
1& Drinking/eating alone& 0.9888\\ \hline
\cellcolor{green!25}2& \cellcolor{green!25}Reading& \cellcolor{green!25}0.0112\\ \hline
3& Socializing& 0.0000\\ \hline
4& Public Transport& 0.0000\\ \hline
5& Eating together& 0.0000\\ \hline
\end{tabular}
}
\end{minipage}
\end{minipage}\hspace{\spaceBetweenColumns}
\begin{minipage}{.2\textwidth}
\centering
\begin{minipage}{\columnProportion\textwidth}
\centering
{\helvetica Shopping}
\includegraphics[scale=\imgscale]{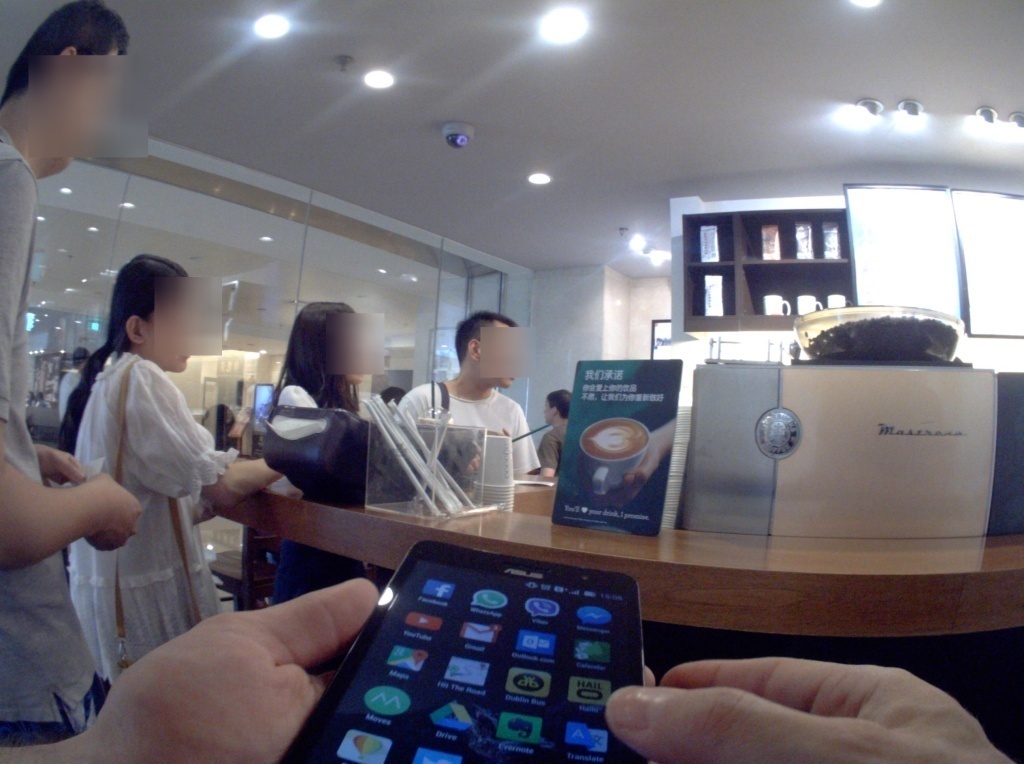}
\end{minipage}\vspace{\spaceBetweenBoxes}

\begin{minipage}{\columnProportion\textwidth}
\centering
{\helvetica VGG-16 Top 5}
\resizebox{\boxProportion\columnwidth}{!}{%
\begin{tabular}{|c|c|c|}
\hline
\textbf{\#} & \textbf{Activity} & \textbf{Score} \\ \hline
1&Mobile&0.6543\\ \hline
2&Drinking together&0.1352\\ \hline
3&Socializing&0.0967\\ \hline
\cellcolor{green!25}4&\cellcolor{green!25}Shopping&\cellcolor{green!25}0.0912\\ \hline
5&Attending a seminar&0.0095\\ \hline
\end{tabular}
}
\end{minipage}\vspace{\spaceBetweenBoxes}

\begin{minipage}{\columnProportion\textwidth}
\centering
{\helvetica ResNet Top 5}
\resizebox{\boxProportion\columnwidth}{!}{%
\begin{tabular}{|c|c|c|}
\hline
\textbf{\#} & \textbf{Activity}  & \textbf{Score} \\ \hline
1&Mobile&0.9953\\ \hline
2&Shopping&0.0020\\ \hline
3&Drinking together&0.0016\\ \hline
4&Socializing&0.0009\\ \hline
5&Meeting&0.0001\\ \hline
\cellcolor{red!25}19&\cellcolor{red!25}Walking outdoor&\cellcolor{red!25}0.0000\\ \hline
\end{tabular}
}
\end{minipage}\vspace{\spaceBetweenBoxes}

\begin{minipage}{\columnProportion\textwidth}
\centering
{\helvetica InceptionV3 Top 5}
\resizebox{\boxProportion\columnwidth}{!}{%
\begin{tabular}{|c|c|c|}
\hline
\textbf{\#} & \textbf{Activity}  & \textbf{Score} \\ \hline
1&Mobile&0.8928\\ \hline
2&Drinking together&0.0291\\ \hline
\cellcolor{green!25}3&\cellcolor{green!25}Shopping&\cellcolor{green!25}0.0285\\ \hline
4&Socializing&0.0135\\ \hline
5&Walking indoor&0.00953\\ \hline
\end{tabular}
}
\end{minipage}
\end{minipage}\hspace{\spaceBetweenColumns}
\begin{minipage}{.2\textwidth}
\centering
\begin{minipage}{\columnProportion\textwidth}
\centering
{\helvetica Plane}
\includegraphics[scale=\imgscale]{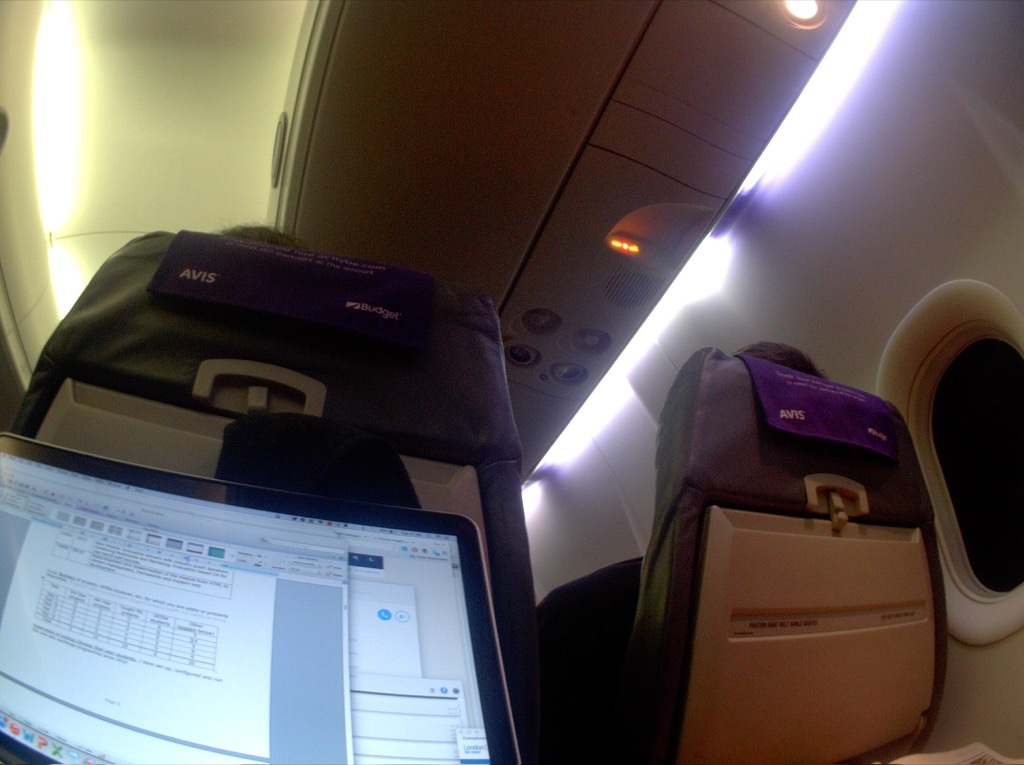}
\end{minipage}\vspace{\spaceBetweenBoxes}

\begin{minipage}{\columnProportion\textwidth}
\centering
{\helvetica VGG-16 Top 5}
\resizebox{\boxProportion\columnwidth}{!}{%
\begin{tabular}{|c|c|c|}
\hline
\textbf{\#} & \textbf{Activity} & \textbf{Score} \\ \hline
1	& Plane	& 0.5362\\ \hline
2	& Reading	& 0.3613\\ \hline
3	& Working	& 0.0602\\ \hline
4	& Attending a seminar	& 0.0236\\ \hline
\cellcolor{red!25}5	& \cellcolor{red!25}Public Transport & \cellcolor{red!25}0.0167\\ \hline
\end{tabular}
}
\end{minipage}\vspace{\spaceBetweenBoxes}

\begin{minipage}{\columnProportion\textwidth}
\centering
{\helvetica ResNet Top 5}
\resizebox{\boxProportion\columnwidth}{!}{%
\begin{tabular}{|c|c|c|}
\hline
\textbf{\#} & \textbf{Activity}  & \textbf{Score} \\ \hline
1	& Plane	& 0.9939\\ \hline
2	& Attending a seminar	& 0.0048\\ \hline
3	& Cooking	& 0.0009\\ \hline
4	& Mobile	& 0.0003\\ \hline
5	& Reading	& 0.00003\\ \hline
\cellcolor{red!25}6 & \cellcolor{red!25}Public Transport & \cellcolor{red!25}0.0000\\ \hline
\end{tabular}
}
\end{minipage}\vspace{\spaceBetweenBoxes}

\begin{minipage}{\columnProportion\textwidth}
\centering
{\helvetica InceptionV3 Top 5}
\resizebox{\boxProportion\columnwidth}{!}{%
\begin{tabular}{|c|c|c|}
\hline
\textbf{\#} & \textbf{Activity}  & \textbf{Score} \\ \hline
1	& Plane	& 0.6574\\ \hline
\cellcolor{red!25}2	& \cellcolor{red!25}Working	& \cellcolor{red!25}0.1944\\ \hline
3	& Attending a seminar	& 0.0433\\ \hline
4	& Mobile	& 0.0253\\ \hline
5	& Reading	& 0.0207\\ \hline
\cellcolor{red!25}6 & \cellcolor{red!25}Public Transport & \cellcolor{red!25}0.0195\\ \hline
\end{tabular}
}
\end{minipage}
\end{minipage}\hspace{\spaceBetweenColumns}
\begin{minipage}{.2\textwidth}
\centering
\begin{minipage}{\columnProportion\textwidth}
\centering
{\helvetica Public Transport}
\includegraphics[scale=\imgscale]{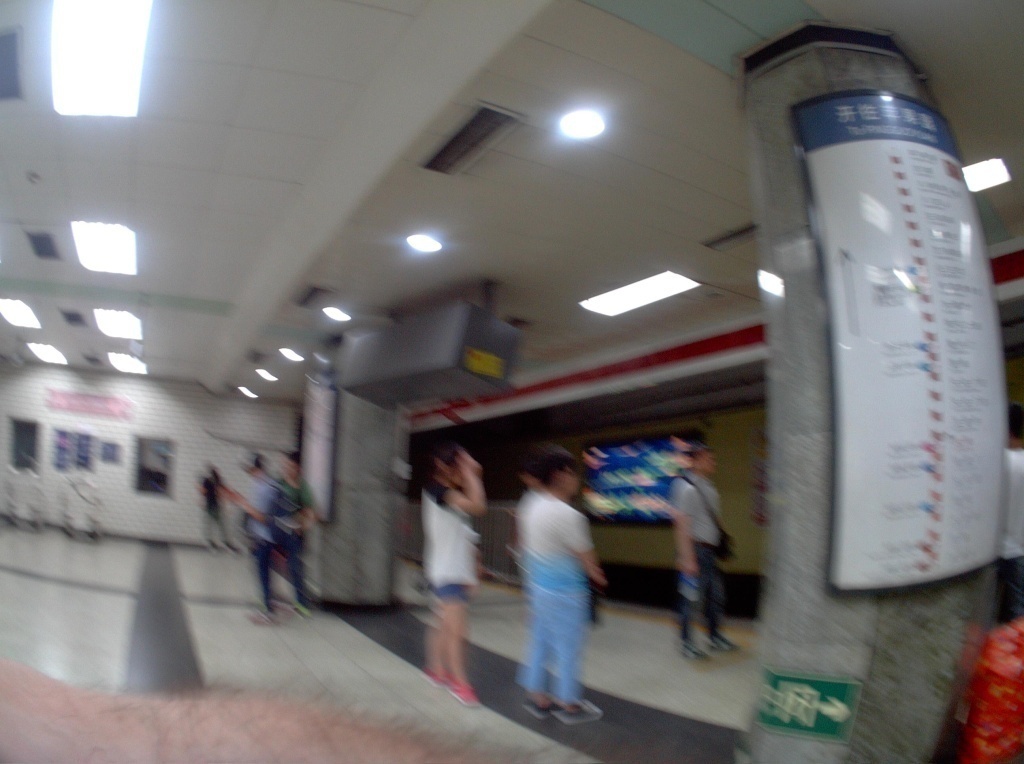}
\end{minipage}\vspace{\spaceBetweenBoxes}

\begin{minipage}{\columnProportion\textwidth}
\centering
{\helvetica VGG-16 Top 5}
\resizebox{\boxProportion\columnwidth}{!}{%
\begin{tabular}{|c|c|c|}
\hline
\textbf{\#} & \textbf{Activity} & \textbf{Score} \\ \hline
1	& Shopping & 0.7789\\ \hline
2 &	Walking indoor & 0.2174\\ \hline
3	& Talking & 0.0022\\ \hline
4	& Mobile & 0.0005\\ \hline
5	& Socializing	& 0.0004\\ \hline
\cellcolor{red!25}6	& \cellcolor{red!25}Walking outdoor & \cellcolor{red!25}0.0003\\ \hline
\end{tabular}
}
\end{minipage}\vspace{\spaceBetweenBoxes}

\begin{minipage}{\columnProportion\textwidth}
\centering
{\helvetica ResNet Top 5}
\resizebox{\boxProportion\columnwidth}{!}{%
\begin{tabular}{|c|c|c|}
\hline
\textbf{\#} & \textbf{Activity}  & \textbf{Score} \\ \hline
1	& Shopping	& 0.9999\\ \hline
2	& Walking indoor	& 0.00006\\ \hline
\cellcolor{red!25}3	& \cellcolor{red!25}Walking outdoor	& \cellcolor{red!25}0.00002\\ \hline
4	& Eating together	& 0.00000\\ \hline
5	& Drinking/eating alone	& 0.0000\\ \hline
\end{tabular}
}
\end{minipage}\vspace{\spaceBetweenBoxes}

\begin{minipage}{\columnProportion\textwidth}
\centering
{\helvetica InceptionV3 Top 5}
\resizebox{\boxProportion\columnwidth}{!}{%
\begin{tabular}{|c|c|c|}
\hline
\textbf{\#} & \textbf{Activity}  & \textbf{Score} \\ \hline
1	& Walking indoor & 0.6220\\ \hline
2	& Shopping	& 0.1909\\ \hline
3	& Mobile	& 0.0415\\ \hline
\cellcolor{red!25}4	& \cellcolor{red!25}Walking outdoor	& \cellcolor{red!25}0.0369\\ \hline
5	& Drinking together	& 0.0157\\ \hline
\end{tabular}
}
\end{minipage}
\end{minipage}

\caption[]{Classification activity examples. On top of each image is shown its true activity label and on bottom its top 5 predictions by VGG-16, ResNet50 and InceptionV3. Additionally, the result of the ensembles \textit{VGG-16+RF on FC1+Softmax}, \textit{ResNet50+RF on Avg. Pooling}, and \textit{InceptionV3+RF on Global Avg. Pooling} is highlighted on color in its corresponding table. The green and red colors means true positive and false positive classification, respectively.}
\label{fig:classification_examples}
\end{figure*}

\textbf{Activity recognition at image level} The results of adding contextual information from fully connected layers are presented on Table \ref{tab:classificationComparison}. They show that CNN LFE improves the performance for all baseline CNN on different ensembles. Moreover, Table \ref{tab:classificationComparison} shows that the best ensembles were the \textit{InceptionV3+RF on Global Avg. Pooling}, \textit{VGG-16+RF on FC2 + Predictions},  and \textit{ResNet+RF on average pooling}. Furthermore, these ensembles improved the baseline accuracy by $8.2\%$, $6.04\%$, and 1.31\% for InceptionV3, VGG-16, and ResNet, respectively. Specifically, the recall of some categories with fewer learning instances improved significantly, such as \textit{Cooking}, and \textit{Cleaning and Choring}. Additionally, high overlapping classes such as \textit{Drinking/Eating alone} and \textit{Eating together} also improved their accuracy. The improvement over the overlapping classes can also be seen on the confusion matrices shown in Fig. \ref{fig:confusion_matrices}. This means that the random forest improved the classification of images belonging to categories that score similar probabilities. Moreover, the only decrease on accuracy is presented on the class \textit{Plane}. Since its accuracy on the baseline CNN is very high ($87.67\%$) considering the small number of learning instances (1,026), we believe this decrease is a consequence of the random forest trying to balance the prediction error among classes. Some classification examples are shown in Fig. \ref{fig:classification_examples}.

\textbf{Activity recognition at batch level} 
The results of using temporal contextual information from batch sequences is shown on Table \ref{tab:performanceSummaryExperiment2}. The fact that accuracy performance is much higher when using the splits defined as in Fig. \ref{fig:datasetSplitExperiment2} (a) for the VGG-16 can be easily understood when looking at consecutive frames in Fig. \ref{fig:examplesFrameSequences}. Specially in the case of static activities such as \textit{reading} or \textit{watching TV}, consecutive frames are very similar and putting some of them in training and other in test is almost equivalent at doing the test including training images, which is unfair.

Comparing the performances of VGG-16+LSTM with VGG-16+RF+LSTM, our experiments demonstrated that the activity scores obtained with CNN LFE are better features for further temporal analysis with respect to features computed by a end-to-end architecture. In both cases, the best performances were achieved by a larger timestep.

\textbf{State of the art comparisons}. In Table \ref{tab:performanceSummaryExperiment2}, we compared the proposed method with the CNN baselines, with our previous method \cite{cartas2017recognizing},  with the work of Castro et al. \cite{castro2015predicting} and with \cite{cartas2017batch}, that exploits temporal information. Our proposed approach \textit{CNN+RF+LSTM} gives the best results for each CNN baseline when using 10 as timestep value. Specifically, the \textit{InceptionV3+RF+LSTM} with a 10 timestep achieved 89.85\% accuracy. It is followed by the  end-to-end \textit{VGG-16+LSTM} proposed in \cite{cartas2017batch} that also takes into account temporal information and achieves approximatively 6\% less in terms of accuracy. On the contrary, with \textit{CNN+RF}, being a many-to-one architecture, the best results are achieved when using a small timestep. We can also observe that, in general, methods that take into account temporal information achieve better performance with respect to methods that act at image level. Among these latter methods, not all the fine-tuned CNNs were improved by all LFE methods used for comparison. On this dataset, the methods \cite{cartas2017recognizing} and \cite{castro2015predicting} are almost equivalent. To explain this result we performed further experiments to investigate the contribution of time metadata and we could double check that indeed they help in improving the performances (\textit{VGG-16 + RF on FC1 + softmax + date$\&$time}), a result that would unlikely be generalizable to a dataset with more individuals having a very different lifestyle. The fact that color histogram as contextual feature leads to slightly better performances might be due to the different light conditions imposed the activity categories.

In this study, we limited our state of the art comparisons to activity recognition methods conceived specifically for egocentric images or photo-streams. Indeed, state of the art activity recognition methods developed for videos often rely on features that cannot be computer in a lifelogging setting. For example, the method proposed by Fathi et al. \cite{fathi2011understanding} relies on a hierarchy between actions and activities, and the sequential order of the former; but, given the frame-rate of photo-streams,  our labels do not consider actions. Another method presented by Fathi et al. \cite{fathi2012learning} relies on the egocentric gaze provided by the camera, which is not available for a chest-mounted camera as the one used for capturing our dataset. In addition, the  classification methods proposed in \cite{Ma_2016_CVPR,Singh_2016_CVPR} depend on motion features that can not be computed on photo-streams because of the very low frame rate. Finally, the temporal pyramid approach proposed in \cite{pirsiavash2012detecting} strongly relies on shot detection which is in general difficult to compute in photo-streams \cite{dimiccoli2016sr}, and highly depends on additional cues based on object detection and human-object interaction.

Although our results  are encouraging, we recognize that assessing the generalization capability of our system  would require a dataset captured by a large number of persons having a wide variety of lifestyles. Given that lifelogging is a relatively new area of research, and that building such a benchmark would be very expensive in terms of human resources and annotation effort, we leaved this for future work.

\section{Conclusion}
\label{sec:conclusion}

In this paper, we proposed a new pipeline for activity recognition from egocentric photo-streams that relies on a two-phases approach. First, a CNN late fusion ensemble classifier, which combines different layers of a CNN through a random forest is used to predict the activity probabilities on each image. Specifically, the random forest takes as input a vector containing the output of the softmax probability layer and a fully connected layer encoding global image features. Second, these vector probabilities are used in batches of temporally adjacent images, to train a many-to-many LSTM. In addition, we extended a subset of the NTCIR-12 egocentric dataset consisting of 44,902 images by annotating it with 21 different activity labels. Experimental results on this subset demonstrated that the proposed approach achieve better performances than a end-to-end architecture and state of the art methods. The proposed method achieved an overall accuracy of $89.85\%$.

\section*{Acknowledgements}

A.C. was supported by a doctoral fellowship from the Mexican Council of Science and Technology (CONACYT) (grant-no. 366596). This work was partially founded by TIN2015-66951-C2, SGR 1219, CERCA, \textit{ICREA Academia'2014} and 20141510 (Marat\'{o} TV3). The funders had no role in the study design, data collection, analysis, and preparation of the manuscript. M.D. is grateful to the NVIDIA donation program for its support with GPU card.

\section*{Notes}

A final version of this manuscript was published in the \textit{Pattern Analysis and Applications} journal under the name \textit{Batch-Based Activity Recognition from Egocentric Photo-Streams Revisited}.

\bibliographystyle{spmpsci}

\end{document}